\pdfoutput=1

\documentclass[preprint,12pt]{elsarticle}

\usepackage{amsmath,amsfonts,amssymb}

\usepackage{algpseudocode}
\usepackage{algorithm}
\usepackage{array}
\usepackage{textcomp}
\usepackage{stfloats}
\usepackage{url}
\usepackage{xcolor}
\usepackage{tabularx}
\usepackage{verbatim}
\usepackage{adjustbox}
\usepackage{graphicx}
\usepackage{booktabs}
\usepackage{subcaption}
\usepackage{soul}
\date{}

\usepackage{dsfont}

\usepackage{hyperref}
\hypersetup{
    colorlinks=true,
    linkcolor=blue,
    citecolor=blue,
    filecolor=magenta,
    urlcolor=cyan,
}

\begin{document}

\maketitle

\begin{frontmatter}



\title{Semantic Prioritization in Visual Counterfactual Explanations \\with Weighted Segmentation and Auto-Adaptive Region Selection}

\author[label1]{Lintong Zhang}
\author[label1]{Kang Yin}
\author[label1]{Seong-Whan Lee}
\affiliation[label1]{
            addressline={Department of Artificial Intelligence, Korea University},
            postcode={02841},
            city={Seoul},
            country={Korea}}

\thanks{This work was supported by Institute of Information & communications Technology Planning & Evaluation (IITP) grant funded by the Korea government(MSIT) (No.2022-0-00984,Development of Artificial Intelligence Technology for Personalized Plug-and-Play Explanation and Verification of Explanation).\textit{(Corresponding author: Seong-Whan Lee.)}}
\thanks{Lintong Zhang, Kang Yin and Seong-Whan Lee are with the Department of Artificial Intelligence,
Korea University, Seoul 02841, South Korea (email: \{zhanglintong, charles\_kang, hayoungjo, sw.lee\}@korea.ac.kr).}

\begin{abstract}
In the domain of non-generative visual counterfactual explanations (CE), traditional techniques frequently involve the substitution of sections within a query image with corresponding sections from distractor images. Such methods have historically overlooked the semantic relevance of the replacement regions to the target object, thereby impairing the model's interpretability and hindering the editing workflow. Addressing these challenges, the present study introduces an innovative methodology named as Weighted Semantic Map with Auto-adaptive Candidate Editing Network (WSAE-Net). Characterized by two significant advancements: the determination of an weighted semantic map and the auto-adaptive candidate editing sequence. First, the generation of the weighted semantic map is designed to maximize the reduction of non-semantic feature units that need to be computed, thereby optimizing computational efficiency. Second, the auto-adaptive candidate editing sequences are designed to determine the optimal computational order among the feature units to be processed, thereby ensuring the efficient generation of counterfactuals while maintaining the semantic relevance of the replacement feature units to the target object. Through comprehensive experimentation, our methodology demonstrates superior performance, contributing to a more lucid and in-depth understanding of visual counterfactual explanations.

\end{abstract}

\begin{graphicalabstract}
\begin{figure*}[htb]
    \centering
    \includegraphics[width=\textwidth]{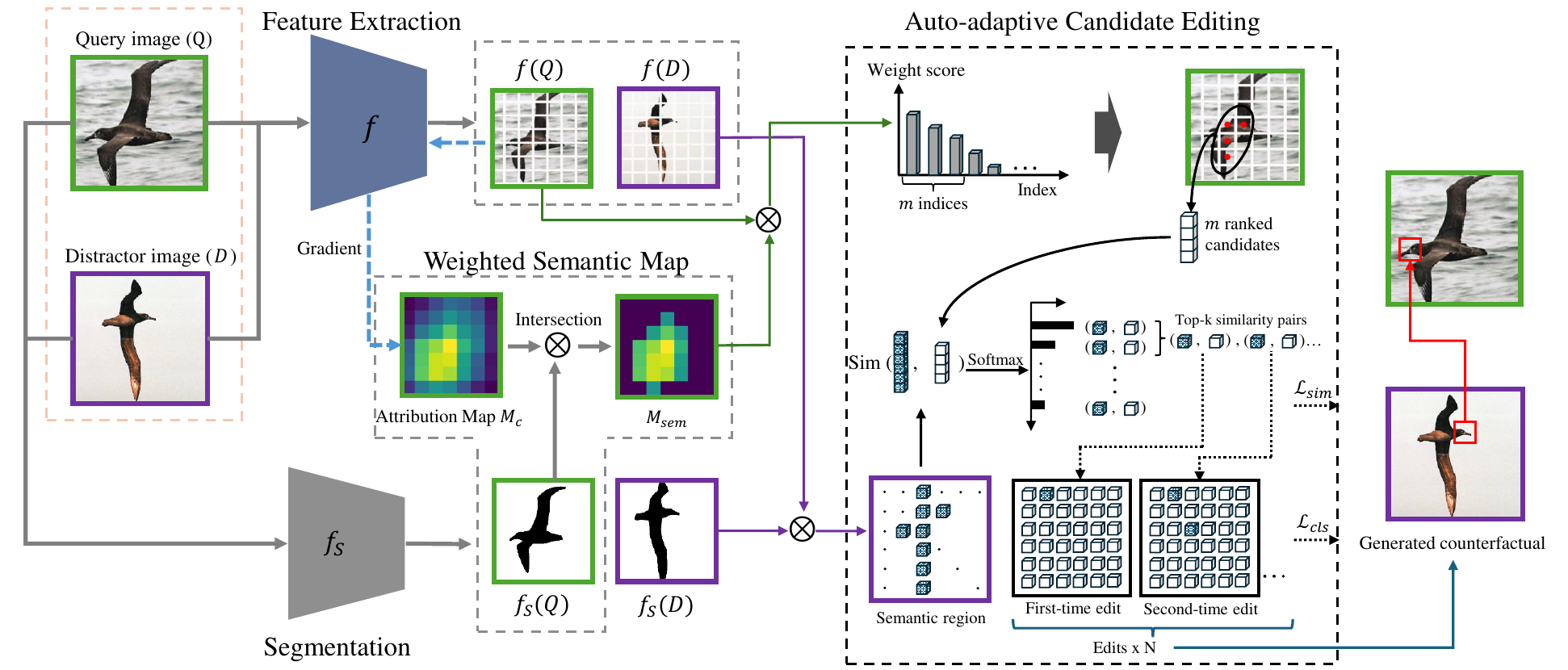}
\end{figure*}
\end{graphicalabstract}

\begin{highlights}
\item The weighted semantic map maximizes constraints on the generation of non-semantically relevant counterfactuals.

\item The efficiency of counterfactual explanation generation is significantly improved through auto-adaptive region selection.
\end{highlights}


\end{frontmatter}


\section{Introduction}
As methods in Explainable Artificial Intelligence (XAI) continue to evolve, particularly within the domain of image classification, numerous studies on interpretability have emerged. These include feature attribution-based explanations~\cite{1,2} and saliency maps~\cite{3}. The~\cite{4} introduced an instance-specific explanation, specifically counterfactual explanations (CEs). Moreover, a method based on visualizing neurons~\cite{5} elucidates the category information contained within neural network neurons. However, when explanatory methods are applied to fine-grained image classification tasks, the subtle differences in features between categories can lead to biases in the interpreted feature regions.

For fine-grained classification tasks, prototype based gray-box models, e.g., prototypical part network (ProtoPNet) method~\cite{6}, region grouping for feature-based method~\cite{7}. While these methods are capable of identifying which features are crucial for model predictions, they fail to specify how these features could be altered to change the predicted outcomes. In fine-grained classification tasks, the presence of multiple similar classes often leads saliency-based explanation methods~\cite{1,2,3} to highlight similar regions, making it challenging to interpret class decisions between categories with subtle differences. By identifying discriminative regions using counterfactual methods, it becomes possible to explain which specific local features the model leverages to differentiate between similar categories. This approach effectively aids users in understanding the subtle distinctions between categories by focusing on these discriminative regions. Counterfactual explanations provide a direct modification of conditions to illustrate the resulting changes, thereby establishing a clear causal relationship. Compared to other explanation methods, such as weight distributions or importance scores, counterfactual explanations are often more intuitive and accessible for non-expert users.



\begin{figure*}[htbp]
  \centering
  \begin{subfigure}[b]{0.41\textwidth} 
    \centering
    \includegraphics[width=\textwidth]{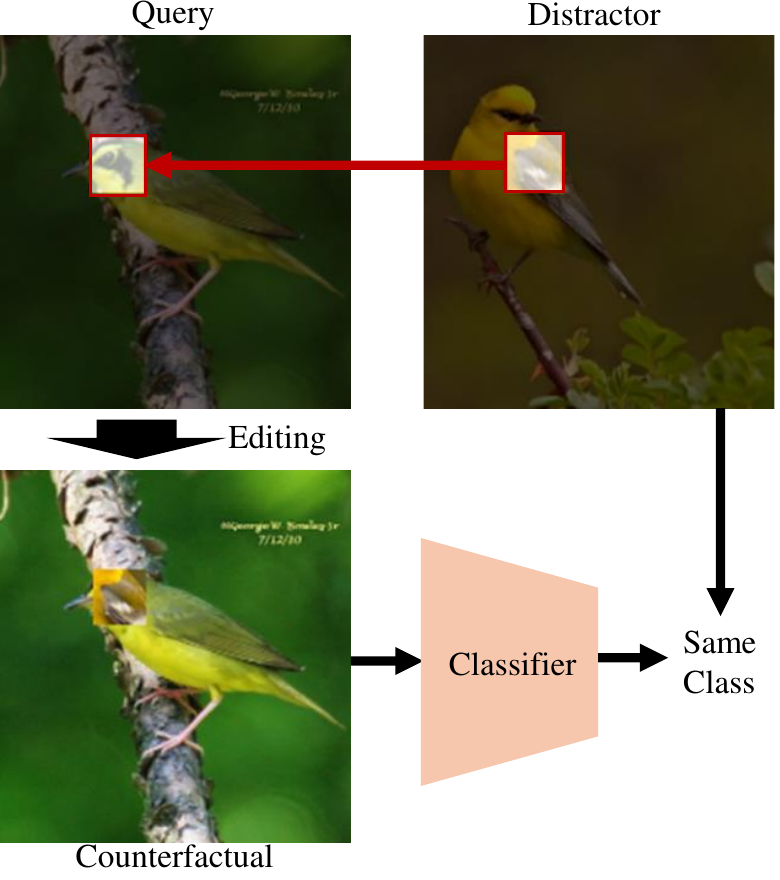} %
    \caption{Counterfactual editing}
    \label{fig:subfig1a}
  \end{subfigure}
  \hspace{0.02\textwidth} 
  \begin{subfigure}[b]{0.50\textwidth} 
    \includegraphics[width=\textwidth]{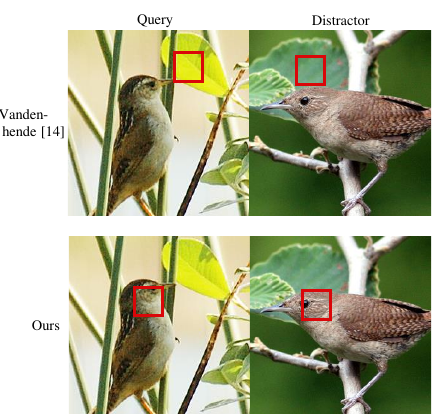}
    \caption{Semantic relevance to the target object}
    \label{fig:subfig1b}
  \end{subfigure}
  \caption{(a) Counterfactual visual explanations method based on region-editing replacement. (b) Compared with previous work~\cite{14}, our method demonstrates greater semantic relevance to the target object.}
  \label{fig1}
\end{figure*}

Fine-grained visual Counterfactual Explanations can be broadly categorized into two streams: generative model-based and non-generative model-based approaches. Generative model-based counterfactual explanation approaches~\cite{8,9,10} utilize the capabilities of generative models to create alternative versions of the input image by modifying specific features associated with different classes. While this method provides a powerful means to explore a diverse range of counterfactual scenarios, it carries the risk of merging features from distinct classes, potentially leading to the generation of counterfactuals that are not only unpredictable but may also compromise the reliability of the explanations. Such outcomes may result in images that deviate significantly from plausible or coherent alterations of the original image.

On the other hand, non-generative approaches include attribution map-based research, as discussed in~\cite{11}, focusing on computing decision regions for counterfactual categories. Additionally, another method employs saliency map-based counterfactual contrasts~\cite{12}, using comparisons between the positive and negative maps of two categories to elucidate differences. These methods emphasize refining highlighted regions within attribution maps to identify the most salient features of an image. However, an exclusive reliance on the regions delineated by attribution maps can lead to counterfactuals exhibiting semantic inconsistencies, as the derived explanations might not fully align with the original semantic context of the image.

A specific category of counterfactual research based feature replacement~\cite{13,14} have proposed a method for generating counterfactual by substituting regions of the query image with those from distractor images, as illustrated in Fig.~\ref{fig1} (a). This method stands out by offering visualizable counterfactual outcomes without necessitating the prolonged durations associated with generative processes. Nonetheless, this method has its flaws, especially with semantic inconsistencies in the replacement regions. Specifically, the approach delineated in~\cite{13}, which involves the replacement of query image regions with those from a single distractor image, reduces the diversity of potentially selectable regions.  The decrease in diversity produces counterfactuals that are challenging for humans to understand and limits the choice of edit combinations that preserve semantic similarity or consistency. Addressing these shortcomings, subsequent research presented in~\cite{14} advocates for the incorporation of multiple distractor images. This enhancement broadens the spectrum of selectable regions, thus mitigating the diversity constraints previously noted in~\cite{13} and enhancing robustness of the method. For semantic consistency,  they introduced a self-supervised model to conduct additional feature extraction and compute semantically similar units, thereby obtaining semantically consistent counterfactuals. However, this method only considers the semantic similarity of features without assessing whether the selected features are relevant to the target object. As illustrated in Fig.~\ref{fig1}(b), studies such as those by~\cite{14} have documented instances where feature regions chosen from the query and distractor images, although semantically similar but the feature region do not relate to the target object. This discrepancy has resulted in counterfactuals that are perplexing to humans. Simultaneously, the replacement of these non-semantic units diminishes the efficiency of counterfactual computation. In contrast, our method confines selections to the semantic scope of the target object, thereby enhancing the comprehensibility of the counterfactuals.

To address the issues outlined previously, we introduce an \textbf{W}eighted \textbf{S}emantic map with \textbf{A}uto-adaptive candidate \textbf{E}diting \textbf{N}etwork (WSAE-Net). Central to our approach is the development of a two-stage editing strategy, consisting of a weighted semantic map generation followed by an auto-adaptive candidate editing sequence. The weighted semantic map is a semantic map that contains information about the importance of feature units during the process of feature unit replacement. To obtain a weighted semantic map, we utilized an attribution explanation method~\cite{15} that identifies decision-contributing regions, and a semantic segmentation method~\cite{16} that delineates semantic scopes. Its objective is to maximize the reduction of the total number of feature units required during the counterfactual computation process, thereby enhancing overall efficiency. The goal is to exclude semantically unimportant feature units from the entire counterfactual computation space, thereby preventing their involvement in the counterfactual computation process. Auto-adaptive candidate editing sequences are designed for more efficient counterfactual editing, deriving an optimal sequence of unit edits from the weighted semantic map based on the importance of units in descending order. Finally, the feature map is computed according to this sequence of units. This method maximizes the reduction of both the time spent editing and the number of edits required to achieve a counterfactual transition, while ensuring that the edited feature units are semantically related to the target object, further enhancing the overall efficiency of counterfactual computation.

Overall, the paper makes three contributions:
\begin{itemize}

    \item we propose an efficient counterfactual computation framework that maximizes the reduction of feature units needing editing and facilitates an easier transition to counterfactual class;

    \item we propose a method for computing counterfactuals that takes into account whether semantics exist for feature units;

    \item we propose a verification method named Average Probability Difference for class transition (APD), determine the relationship with the efficiency of counterfactual generation.

\end{itemize}


\begin{figure}[htbp]
  \centering
  \begin{subfigure}[b]{0.47\textwidth}
    \centering
    \includegraphics[width=\textwidth]{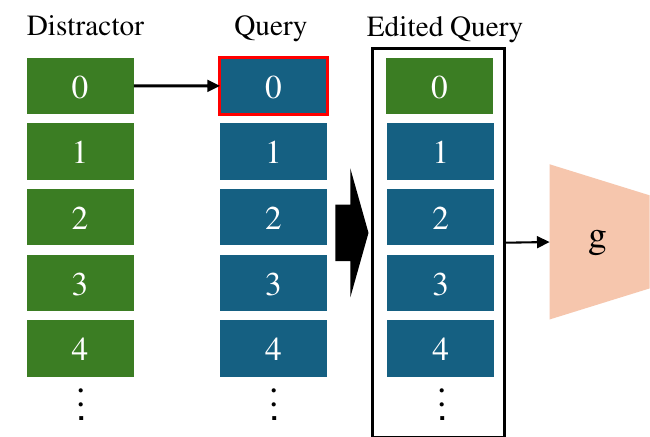}
    \caption{First replacement}
    \label{fig:subfig2a}
  \end{subfigure}
  \hfill 
  \begin{subfigure}[b]{0.47\textwidth}
    \centering
    \includegraphics[width=\textwidth]{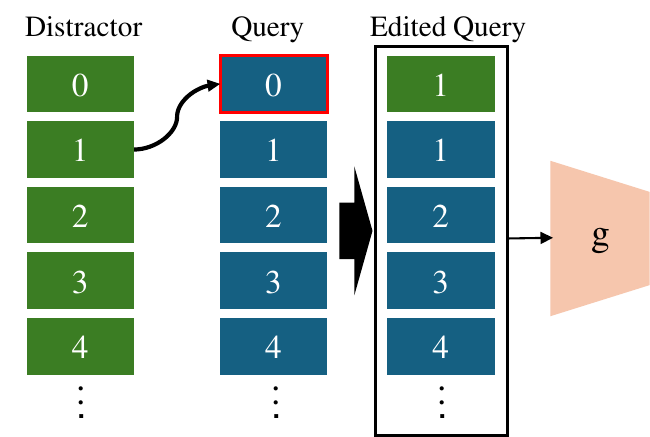}
    \caption{Second replacement}
    \label{fig:subfig2b}
  \end{subfigure}
  \caption{Exhaustive editing of feature units: sequentially replacing all cells of the distractor with the cell at position indices 0 (a) and 1 (b) of the queries and predicted by classifier \textit{g}.}
  \label{fig:sequentially replace all the cells}
\end{figure}

\section{Related Work}
\subsection{Fine-grained Visual Counterfactual Explanation}

The goal of fine-grained visual counterfactual explanations is to find the smallest feature region that leads to a change in decision.  Adversarial attacks~\cite{17,18,19} serve as a notable example, where continuous noise is incrementally added to an image of class $A$ until the classifier's decision changes to class $B$.  Recent advancement in generative method~\cite{20,21,22}, particularly those employing generative models, have shown promise. These advanced methodologie, typically applied to datasets like MNIST~\cite{23} and CelebA~\cite{24}, utilize Generative Adversarial Networks (GANs) to craft more interpretable explanations.  Additionally, in non-generative counterfactual approaches, there is a distinct category of research focusing on the identification and modification of specific concepts or regions within images to alter prediction outcomes. Tools like~\cite{25} utilize attribution maps to pinpoint these critical regions, while approaches like CoCoX~\cite{26} detect visual concepts that, when added or removed, change the model's predictions. Common strategies~\cite{13,14} involve using distractor images $I'$ from class $B$ to determine which parts to replace in the query image $I$ from class $A$, effectively altering the model's prediction to class $B$. This research offers two primary advantages: Firstly, it utilizes distractor images which can be readily sourced from existing datasets; this method requires only images from categories distinct from the query image. Secondly, the approach is well-suited to fine-grained recognition tasks, where alterations in small areas can lead to changes in class assignment. Ultimately, these methods avoid generative models, thereby obviating the need for a learning process and simplifying model explanation.

Although previous methods have seen continuous improvements, the underlying editing strategy and region have remained unchanged. As depicted in Fig.~\ref{fig:sequentially replace all the cells}, the editing region continues to encompass all feature units from position index 0 to the final position after the feature map has been flattened. The strategy initiates with the first unit, sequentially replacing each subsequent unit to compute category predictions. This method still exhibits low computational efficiency; therefore, our work seeks to overcome these limitations by building upon the foundations established by the research institution.

\subsection{Attributive Explanations}

Attributive explanations, as a prevalent approach in visual explanation methodologies, focus on pinpointing the specific pixels or regions within an image that significantly influence a classifier's prediction. The heatmap-based methods~\cite{27,28,29,30} vividly illustrate which areas or features contribute most significantly to the model's decision-making process. Compared to heatmap-based methods, class activation map-based methods are a more refined visualization technique that not only identifies which parts are important but also specifies the precise contribution (positive or negative) of each feature to the model's output. Earlier frameworks primarily focused on gradient-based methods~\cite{31,32,3}, which calculate the gradients of the classifier relative to specific inputs or layers of the network. It is important to note that the attribution maps generated by these methods typically rely on the last convolutional layer, which results in a lower resolution and notably blurred target activation maps. The approach described in~\cite{15} effectively mitigates this issue. 

\subsection{Semantic Segmentation}

In the field of image semantic segmentation, recent advancements have led to the emergence of Swin Transformer~\cite{33}-based semantic segmentation models~\cite{34}, which have gradually achieved top-tier performance. PVT~\cite{35} introduces a progressive shrinking pyramid backbone network that reduces computational costs while providing finer-grained segmentation outputs. Additionally, Mask2Former~\cite{36} presents a novel transformer architecture suitable for various segmentation tasks, including panoptic, instance, and semantic segmentation, marking a significant step towards a universal segmentation framework. Segment Anything (SAM)~\cite{16}, another noteworthy model, exhibits the ability to generate segmentation masks for any dataset based solely on an input image and a corresponding prompt.

\section{Methodology}

\subsection{Problem Formulation}
Let $\mathcal{I}$ be an image set with $\mathcal{C}$ category. Within a category $c, c \in \mathcal{C}$ containing $n_c$ instances, define the $I_i^{c}$ is $i$-th query image for category $c$ and query set as $Q=\{I_i^{c}\}_{i=1}^{n_c}$. Similarly, define the $I_i^{c'}$ is $i$-th image of a category $c', c' \in \mathcal{C}, c'\neq c$. So distractor set is $D=\{I_i^{c'}\}_{i=1}^{n_{c'}}$, ensuring that $Q\cup D=\mathcal{I}$ and $Q\cap D = \varnothing$. Consider $I$ and $I'$ to be randomly chosen samples from $Q$ and $D$, belonging to categories $c$ and $c'$, respectively.

Expanding on the insights of~\cite{13}, we explore a feature extraction function $f:\mathcal{I}\to \mathbb{R}^{HW\times d}$, where $H$ and $W$ denote the spatial dimensions and $d$ the number of channels. In addition, a classifier $g:\mathbb{R}^{HW\times d}\to\mathbb{R}^{\mathcal{C}}$ links the spatial features to a series of $\mathcal{C}$-logits. By replacing certain spatial elements of $f(I)$ with those from $f(I')$, we obtained $f(I^*)$, ultimately leading the classifier to predict category $c'$ for $f(I^*)$. The counterfactual instance $f(I^*)$ can be described by the formula:
\begin{equation}
f(I^*)=(\mathds{1} -\mathbf{a})\circ f(I) + \mathbf{a}\circ Pf(I'),
\end{equation}
where sparse binary gating vector $\mathbf{a}\in \mathbb{R}^{HW\times d}$ modulates the blend between $f(I')$ and $f(I)$ in the final output $f(I^*)$. $P\in \mathbb{R}^{HW\times HW}$ acts as a permutation matrix to align $f(I')$ with $f(I)$, and $\circ$ symbolizes the Hadamard product. The objective is to maximize the counterfactual class probability  $g_{c'}( \cdot)$, thereby facilitating the modification from class $c$ to class $c'$ . The loss function implemented to achieve this goal is:
\begin{equation}
L_{\text{cls}} = \max_{P, a} g_{c'}\left((\mathds{1} - \mathbf{a}) \circ f(I) + \mathbf{a} \circ P f(I')\right),
\label{eq2}
\end{equation}
where $P$ belonging to $\mathcal{P}$, which denotes the set of all permutation matrices of size $HW \times HW$. 

To circumvent the trivial solution where all cells within $I$ are substituted, a sparsity constraint is applied to the variable $\mathbf{a}$ to minimize the quantity of cell edits. Employing the greedy approach in~\cite{13}, we iteratively replace spatial cells within $I$ by iteratively solving for $L_{\text{cls}}$ that maximizes the predicted probability, continuing this process until a shift occurs in the model's decision.

\subsection{Framework Overview}

\begin{figure*}[htb]
    \centering
    \includegraphics[width=\textwidth]{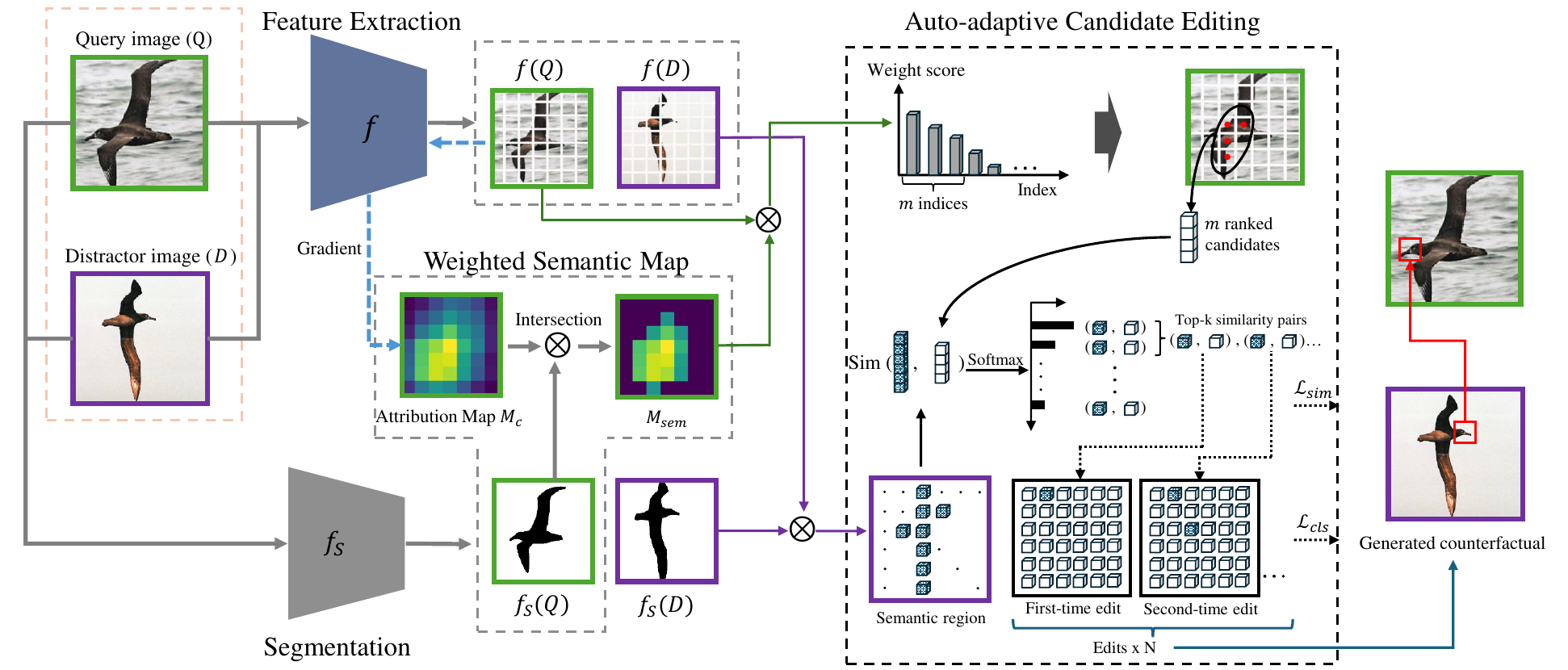}
    \caption{Overview of the proposed WSAR-Net framework. It consists of three key components: feature extraction $f$ and a segmentation model $f_{S}$ to create weighted semantic maps, and sequence optimization during editing. By integrating the model's attribution map with segmentation matrices, we establish a weighted semantic map for the query. Simultaneously, the semantic region is defined based on the segmentation matrix from the distractor image. In the auto-adaptive candidate editing phase, we filter for the top-$m$ units by weight scores in the weighted semantic map, then compute similarity against the semantic features of the distractor image to derive similarity scores. From these, the top-$k$ units with the highest similarity are combined with the query image's $m$ units with the highest weight score. Subsequent replacements in the query's feature map based on these combinations lead to recalculating the counterfactual class until a change in the predicted class transition occurs. $L_\text{sim}$ and $L_\text{cls}$ represent the similarity and classification losses, respectively.}
    \label{fig.framework}
\end{figure*}

The proposed WSAR-Net, as shown in Fig.~\ref{fig.framework}, is designed in two primary phases: one dedicated to generating weighted semantic maps, and subsequently, an auto-adaptive phase for editing candidate sequences. The essence of the weighted semantic map is a matrix comprised solely of class weight values associated with existing semantic units, with the weight values of units lacking semantic content set to zero. During the initial phase, a feature extraction module is employed to derive the latent representations of the query and distractor images, alongside generating an attribution map for the query image through backward class-specific gradients. Concurrently, a segmentation module is utilized to discern the semantic elements within the images. By integrating attribution maps with the segmentation matrix, we have derived a weighted semantic map. The auto-adaptive phase for editing candidate sequences aims to identify an optimal sequence of unit edits within the semantics of the query, ensuring rapid transitions of counterfactual classes and enhancing the efficiency of counterfactual computations. In the auto-adaptive candidate editing phase, the features of the query are ranked according to their significance from the weight score in the weighted semantic map, with the top-$m$ ranked candidates designated for priority computation. For the distractor, calculations are only performed at positions where the corresponding segmentation matrix values are non-zero. Subsequently, we compute the similarity for the feature units corresponding to the query and distractor, obtaining a similarity score. Ultimately, we calculate only the top-$k$ similarity-scored editing combinations until a class transition is predicted.

\subsection{Weighted Semantic Map Generation}

We follow attribution function to get the attribution map of query image. Given a query image $q$ from $Q$ with class $c$, the single attribution map $M_c$ is formulated as follows:

\begin{equation}
\begin{aligned}
M_c &= \phi\left( \sum_k \phi(b^k_c) \cdot A^k \right),
\end{aligned}
\label{eq3}
\end{equation}
where $b^k_c$ is attribution for class $c$ and $A^k$ is the $k^{th}$ feature map, and $\phi$ denotes the ReLU activation function. We then use $M_c$ as the importance weights of the following segmentation part.

For a given images $\mathcal{Z}=(q, d)$, $q,d$ are single query and distractor images, we then apply segmentation model $f_S$ to get segmentation matrix of semantics from the background, and the model outputs a binary matrix indicating the location of semantics for $q$ and $d$ each, i.e., $(S_q, S_d)= f_S(\mathcal{Z})$, which is identical to the input in shape.  Afterwards, we apply bilinear interpolation to reduce the size of segmentation matrix to match attribution map for further operations.

Different from traditional approaches that directly use $M_c$ to assign feature importance, we innovatively combine it with the segmentation matrix. Due to the fact that much of the feature units are actually semantically meaningless, it is unhelpful to consider those areas for editing. Therefore, we naturally take the segmentation output $S_q$ as the semantic threshold, and generate weighted semantic matrix $M_{sem}$ as follows: 
\begin{equation}
M_{sem} = S_q \circ M_c,
\label{eq4}
\end{equation}
where $\circ$ is Hadamard product. It should be noted that $M_{sem}$ and $M_c$ are computed only for the query image $q$, but not for the distractor $d$.

The matrix $M_{sem}$ serves as a refined attribution map, highlighting areas critical to the classifier’s decisions and sidelining less relevant units. By applying this method, we limit the editing region to exclude non-target units, addressing a common shortfall in prior studies. This targeted selection ensures that only pertinent feature units within the object's region are considered for subsequent replacements and predictions on the feature map.

\subsection{Auto-adaptive Candidate Editing Process}
\subsubsection{Optimal editing sequence} 
To enhance the editing strategy, during the editing process, we first compute only those editing combinations that are semantically relevant to the target object from the total number of editing combinations. Second, we aim to obtain the optimal sequence for computing these editing combinations to ensure rapid transitions in counterfactual class.

Firstly, it is essential to significantly reduce the overall number of edit combinations required for computation. The total number of edit combinations for a query and distractor images is $T_E = HW\times HW$ as in~\cite{13,14}, where the two $HW$ represents the length of flattened feature maps of $f(q)$ and $f(d)$, respectively. Hence, the whole edit combination $E = \{(i,j)~|~i, j\in \{1,2,...,HW\}\}$, where $i$ and $j$ are the corresponding position indices in $f(q)$ and $f(d)$. 

Unlike previous methods, we carefully select a part of edit combinations based on non-zero values in segmentation matrix $S_q$ and $S_d$ and define $E_s$ as the selected editing combination. Therefore, the selected total number of edit combinations $T_{E_s}$ for a query and distractor images are represented as:
\begin{equation}
T_{E_s} = \underbrace{\left( \sum_{i=1}^{HW}  \mathds{1}_{\{S_{q_i}\neq 0\}} \right)}_{N_Q} \times \underbrace{\left(  \sum_{j=1}^{HW} \mathds{1}_{\{ S_{d_j}\neq 0\}} \right)}_{N_D}~,
\end{equation}
where $S_{q_i}$ and $S_{d_j}$ are the $i$-th, $j$-th units in $S_q$ and $S_d$. The $N_Q$ and $N_D$ are the numbers of non-zero units in the segmentation matrix of query and distractor, respectively. Through this method, we maximize the reduction of computations for editing combinations that lack semantic information.

Secondly, inspired by works~\cite{37,38}, to prioritize the computation of editing combinations that facilitate the transition to a counterfactual class, we have flattened the matrix $M_{sem}$ to facilitate the calculation of the corresponding class weight score for each element. So the calculation process for the weight score of each element $w_i$ in $M_{sem}$ can be represented as:
\begin{equation}
\sigma(w)_i = \frac{\mathds{1}_{\{w_{i}\neq 0\}}e^{w_{i}}}{\sum_{k=1}^{HW} \mathds{1}_{\{w_{k}\neq 0\}} e^{w_{k}}},
\label{eq8}
\end{equation}
where $\sigma$ is the softmax function, subsequently, we can get the list $l$ of weight scores and indices of all elements:

\begin{equation}
l = \left\{ (\sigma(w)_i, i) \,|\, w_i \in M_{\text{sem}}, w_i \neq 0 \right\}.
\label{eq8}
\end{equation}

With a defined weight score threshold $t$, we select the position indices of the $m$ feature units based on their weight scores after ranking, thus establishing the editing sequence for the query feature map. Simultaneously, the editing sequence for the distractor feature map is determined by the sequence of position indices of non-zero units within the associated segmentation matrix. This calculation process for new position sequence list can be expressed as:

\begin{equation}
\begin{aligned}
    l_Q &= \Biggl\{ m \, \Bigg| \, \left(t - \sum_{i=1}^{m} R([l])_i \right) \left(t - \sum_{i=1}^{m+1} R([l])_i \right) < 0, m \in l \Biggr\},\\
    l_D &=\{j\,|\, S_{d_j} \neq 0 , j\in \{1,2,...,HW\}\},
\end{aligned}
\label{eq10}
\end{equation}
where $l_Q$ is the new position sequence list formed after screening based on threshold $t$ for feature map of query, and $R(\cdot)$ is the ranking function with a descending order for the weight scores of list $l$.  $l_D$ is new position sequence in semantic region for feature map of distractor. To guarantee that only the $m$-th weight score is selected in calculation process for new position sequence list, we propose an optimization selection function $L_{\text{o}}$ can be expressed as follows: 
\begin{equation}
L_{\text{o}}= \max\{t-\sum_i^m R([l])_i, 0\}.
\label{eq9}
\end{equation}

\subsubsection{Optimizing Calculations for Multiple Distractors}

The previous work~\cite{14} notes that counterfactual editing based on multiple distractor images results in diversified editing regions that maintain semantic consistency, thereby yielding counterfactuals that are more easily understood by humans. However, this approach also leads to an exponential increase in computational costs with the addition of distractor images.

To manage and optimize the computational burden, according to Eq.~\ref{eq10}, when editing a counterfactual based on a single distractor image, each selected editing combination can be represented as $E_s = [(i,j)~|~i\in l_Q, j\in l_D]$. Therefore, when editing counterfactuals based on multiple distractor images, only the selected editing combinations are computed. Consequently, the total number of editing combinations for counterfactuals based on $n$ distractor images can be expressed as $n \times T_{E_s}$. To preserve the advantages of semantic consistency identified in previous work while further filtering and computing only the editing combinations within $n \times T_{E_s}$ that possess higher semantic similarity, our method is akin to the approach described in~\cite{14}. We estimate a probability distribution for a given query unit $i$ across all distractor units $j'$ using a non-parametric softmax function, which identifies which distractor units are most likely to contain regions semantically similar to the query unit $i$.
Therefore, we calculate the likelihood function $L$ that unit $i$ of $q$ semantically corresponds to cell $j$ of multiple distractor images $\tilde{d}$ by:

\begin{equation}
\begin{aligned}
L &= \log\left(\frac{\exp(f(q)_i \cdot f(\tilde{d})_{j} / \tau)}{\sum_{j' \in f(\tilde{d})} \exp(f(q)_i \cdot f(\tilde{d})_{j'} / \tau)}\right),
\end{aligned}
\label{eq11}
\end{equation}
where $\tau$ is a temperature hyperparameter that relaxes the dot product. And similarity loss function $L_{\text{sim}}$ for this process can be expressed as:

\begin{equation}
L_{\text{sim}}= L(a^T f(q), a^T P f(\tilde{d})),
\label{eq12}
\end{equation}
where the similarity loss $L_{\text{sim}}$ assesses the semantic similarity between selected units in the query image (specified by $\mathbf{a}^T f(q)$ ) and the corresponding units in the distractor image (specified by $\mathbf{a}^T \mathbf{P}f(\tilde{d})$). In order to optimize computational resources, we initially compute the semantic similarity through $L_{\text{sim}}$ to select the top-$k$ (similarity threshold $u$) of unit permutations with the least loss, thereby excluding permutations involving units with dissimilar semantics. Subsequently, we calculate $L_{\text{cls}}$ exclusively on these preselected top-$k$ permutations.

In summary, according to Eq.~\ref{eq2}, and Eq.~\ref{eq12}, the total loss function for computing counterfactuals based on multiple distractor images of the proposed WSAR-Net method is as follows:

\begin{equation}
L_{\text{total}} = L_{\text{cls}} + \lambda L_{\text{sim}},
\end{equation}
where $\lambda$ are hyperparameters to balance the loss.

\section{Experiment}
\subsection{Experiment Setup}

\subsubsection{Datasets} The experiments were carried out using the CUB-200-2011~\cite{39} and Stanford Dogs~\cite{40}.  As illustrated in Table~\ref{dataset}, the CUB-200-2011 dataset is an intensively annotated, fine-grained dataset of bird species, encompassing 200 bird types.  Attributes were defined and assigned to each part based on the bird, and the dataset contained 11,988 images, of which 5,994 images were split into a training set and 5,794 images were split into a validation set. 

\begin{table}[h!]
\centering
\caption{Datasets overview.}
\label{dataset}
\resizebox{0.7\linewidth}{!}{
\begin{tabular}{lccccc} 
\toprule
Dataset & \multicolumn{3}{c}{Statistical results} & \multicolumn{2}{c}{Top-1 Accuracy} \\
\cmidrule(lr){2-4} \cmidrule(lr){5-6} 
       & \#Class & \#Train & \#Val & VGG-16 & Res-50 \\ 
\midrule
CUB-200-2011 & 200    & 5,994 & 5,794 & 81.5   & 82.0   \\
Stanf. Dogs  & 120    & 12,000  & 8,580 & 86.7   & 88.4   \\
\bottomrule
\end{tabular}
}
\end{table}

Stanford Dogs contains images of dogs annotated with keypoint~\cite{41} locations of 24 parts, and the dataset contained 20,580 images. Of these, 12,000 images were split into a training set and 8,580 images were split into a validation set.

\subsubsection{Implementation}  Our method's generality was evaluated on the ResNet-50~\cite{42} and VGG-16~\cite{43} backbone and set~\cite{13} as baseline method. We divided the two networks into components $f$ and $g$ in the final down-sampling layer at $conv5\_1$ in ResNet-50 and $max\_pooling2d\_5$ in VGG-16. We set the batch size as 64 , the number of distractor images required for each counterfactual edit is 20. 

We chose 0.1 for $\lambda$ in the semantic constraint loss function, with the similarity threshold $u$ and weight score threshold $t$ set to 0.2 and 0.5, and $\tau$ set to 0.1. The attribution maps were generated using the attribution function LayerCAM~\cite{15}, capturing gradients from the last down sampling layer at $conv5\_1$ in ResNet-50.
To obtain segmentation maps for the CUB-200-2011 and Stanford Dogs datasets, we employed the SAM function~\cite{16}, inputting images along with keypoints as prompts included in the datasets' attribute files. All experiments were conducted on an NVIDIA RTX A5000 GPU with 24-GB memory. In our experiments, the segmentation model was not part of the overall framework but was used solely as a data preprocessing method.

   

\begin{figure*}[htbp]
  \centering
  \begin{subfigure}[b]{\textwidth}  
    \centering
    \includegraphics[width=\textwidth]{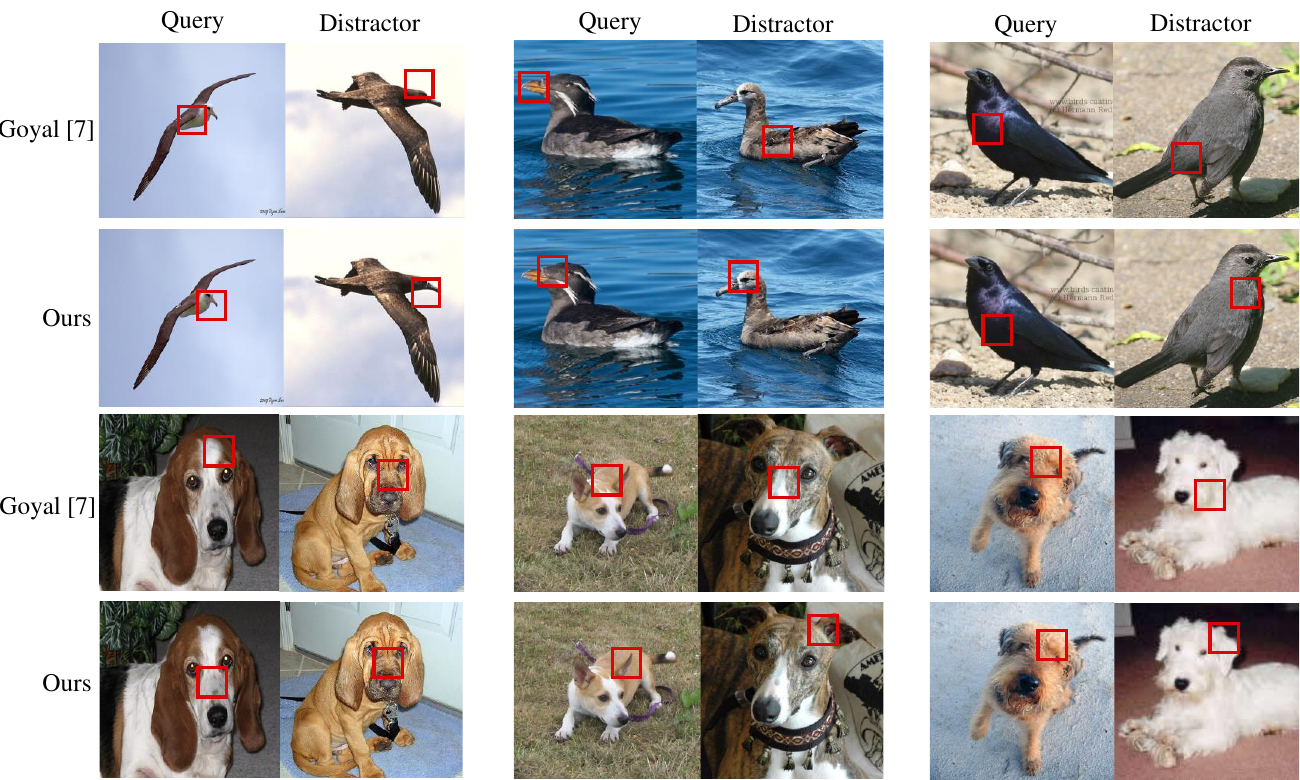}
    \caption{Computing counterfactuals based on single distractor image}
    \label{fig:subfig4a}
  \end{subfigure}
  \hfill 
  \begin{subfigure}[b]{\textwidth}
    \centering
    \includegraphics[width=\textwidth]{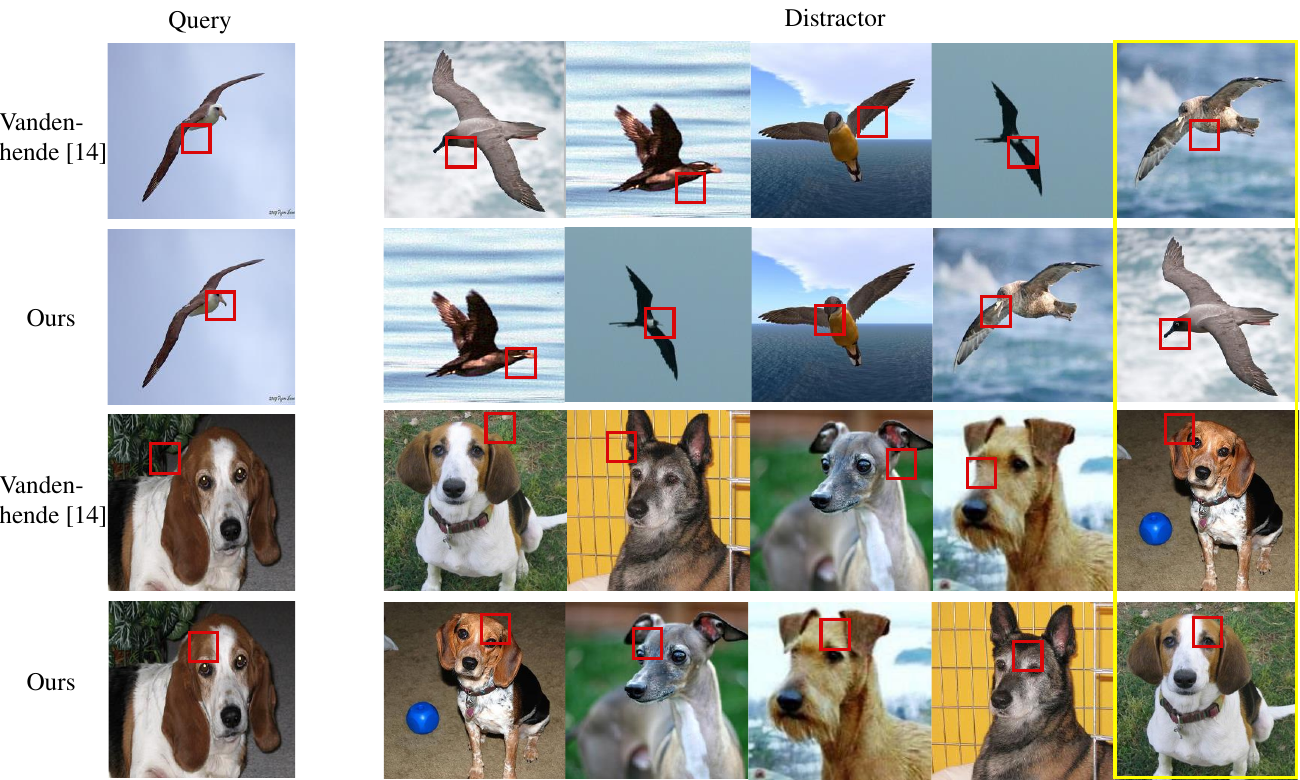}
    \caption{Computing counterfactuals based on multiple distractor images}
    \label{fig:subfig4b}
  \end{subfigure}
  \caption{Visualization of computed counterfactuals. (a) Each query image computes the counterfactual using the feature region from a single distractor image. The area outlined in red represents the region whose replacement into the query image results in a class transition. (b) In the process of selecting units from multiple distractor images for continuous replacement, the distractor image within the yellow area eventually becomes the one containing the replaced unit that leads to the query class transition.}
  \label{app1}
\end{figure*}

\subsection{Evaluation metrics}

We used the following evaluation metrics that were initially used in previous studies~\cite{13,14}:
\begin{itemize}
    \item \textbf{Near-KeyPoint (Near-KP)}. It measures the frequency with which the edited unit on the feature map of the query image and the edited unit on the feature map of the distractor image both contain a keypoint. 

    \item \textbf{Same-KeyPoint (Same-KP)}. It measures how often we select the same keypoints in the feature units of query and feature units distractor image, thus measures semantic consistency of counterfactuals.  

    \item \textbf{\# Edits}. It measures the average number of edits required for the classification model to alter its prediction from the query class to the distractor class across all edited images.
\end{itemize}

It is important to note that evaluating a single edit differs from evaluating all edits because, in the former, we do not assess if the prediction shifts to the counterfactual class. Rather, we only focus on whether the region of the first edit contains semantics or whether  these semantics are similar. For all edits, we continue until the prediction shifts to the counterfactual class, and then evaluate the best edit region identified up to that point.

\begin{table*}[h]
\centering
\caption{Comparison of generated counterfactual using ResNet-50 model and VGG-16 model on CUB-200-2011 and Stanford Dogs datasets, with the Same-KP and Near-KP metrics expressed as (\%).}
\resizebox{\textwidth}{!}{
\begin{tabular}{l ccc ccc ccc ccc}
\toprule
& \multicolumn{6}{c}{CUB-200-2011} & \multicolumn{6}{c}{Stanford Dogs Extra} \\
\cmidrule(lr){2-7} \cmidrule(lr){8-13}
& \multicolumn{3}{c}{ResNet-50} & \multicolumn{3}{c}{VGG-16} & \multicolumn{3}{c}{ResNet-50} & \multicolumn{3}{c}{VGG-16} \\
\cmidrule(lr){2-4} \cmidrule(lr){5-7} \cmidrule(lr){8-10} \cmidrule(lr){11-13}

\multicolumn{13}{c}{\textbf{~~~~~~~~~~~~~~~~~~~~~~~~~~~~~~~~~Single edit}} \\
\midrule
Method & Near-KP & Same-KP & \# Edits & Near-KP & Same-KP & \# Edits  & Near-KP & Same-KP & \# Edits  & Near-KP & Same-KP & \# Edits \\
\midrule
Goyal~\cite{13}      & 61.4 & 11.5 & -   & 67.8 & 17.2 & -  & 42.7 & 6.4 & -    & 42.6 & 6.8 & - \\
Vandenhende~\cite{14} & 71.7 & \textbf{36.1} & -  & \underline{73.5} & \textbf{39.6} & - & \underline{51.2} & \underline{22.6} & -  & \underline{49.8} & \textbf{23.5} & - \\
SCOUT~\cite{11}  & 43.0 & 4.4 & -  & 68.1 & 18.1 & -  & 35.3 & 3.1 & -  & 41.7 & 5.5 & -  \\
Dervakos~\cite{44}  & \underline{65.4} & 20.3 & -  & 69.1 & 26.5 & -  & 49.2 & 18.6 & -  & 48.1 & 17.2 & -  \\
\textbf{Ours}  & \textbf{74.9} & \underline{29.4} & -  & \textbf{75.2} & \underline{38.4} & -  & \textbf{54.3}  & \textbf{23.2 } & -  & \textbf{51.2}   & \underline{22.9} & -  \\

\cmidrule{1-13}
\multicolumn{13}{c}{\textbf{~~~~~~~~~~~~~~~~~~~~~~~~~~~~~~~~~All edits}} \\
\midrule
Goyal~\cite{13}       & 50.9 & ~6.8 & 3.5 & 54.6 & 8.3 & 5.5  & 34.9 & 3.6 &\underline{4.3}  & 35.7 & 3.7 & \underline{6.3} \\
Vandenhende~\cite{14} & \underline{60.3} & \textbf{30.2} & \underline{3.2}  & \underline{68.5} & \underline{35.3} & \underline{3.9}  & \underline{37.2} &\underline{16.7} & 4.8 & \underline{37.5} & \underline{16.4} & 6.6  \\
Dervakos~\cite{44}  & 55.6 & 15.9 & 3.4  & 57.1 & 20.3 & 4.0  & 35.3 & 12.2 & 4.4  & 36.8 &13.9 & 6.4 \\
\textbf{Ours}       & \textbf{61.9} & \underline{29.1} & \textbf{3.0}  & \textbf{70.3} & \textbf{36.5} & \textbf{3.8}  & \textbf{38.1}    & \textbf{ 18.2 } &   \textbf{4.2}   &  \textbf{39.3}  & \textbf{ 17.5 } & \textbf{ 5.9 }   \\
\bottomrule
\end{tabular}
}
\label{table1}
\end{table*}

\subsection{Experiment Result}
\subsubsection{Comparison with state-of-the-art methods} In Table.~\ref{table1}, we compare our method with competitors on CUB-200-2011 dataset and Stanford Dogs dataset with ResNet-50 and VGG-16 model for single edit and all edits. 

First, we conducted a performance evaluation under the single-edit setting. In this setting, class transitions are disregarded, and the focus is solely on the feature region selected during the first iteration. At this stage, our method is primarily limited to selecting the most important feature region for the relevant class, which results in some improvement in the Near-KP metric. This is because the key decision regions for most images generally include KeyPoints. However, the performance decline in the Same-KP metric arises from the fact that the initially selected regions are critical for both the query and distractor classes. The inherent differences between these classes lead to distinct important regions, thereby reducing the probability of selecting regions with the same KeyPoints. As a result, our method shows minimal performance difference compared to method~\cite{14} in the single-edit setting.
Moreover, since our approach primarily addresses semantic relevance and efficiency issues in the successful generation of counterfactuals, the performance differences are mainly observed under the All-edit setting.

By contrast, in our method, the Near-KP metric improved by 10.0\% over~\cite{13}, 1.6\% over~\cite{14} with ResNet-50 on CUB-200-2011 dataset, so the average number of edits have reduced from 3.5 to 3.0. A notable aspect of this comparison is SCOUT~\cite{11}, which diverges from exhaustive search mechanisms for identifying counterfactual edits and resorts to visualization based on a single counterfactual region derived from dual-class attribution maps. This approach limits SCOUT to a singular edit per counterfactual, precluding its comparability in multi-edit scenarios. Additionally, method~\cite{44} primarily excels in retrieving the optimal counterfactual image (distractor image), which generally results in better performance compared to the baseline method. However, this method does not take into account semantic constraints or the optimization of the editing sequence when editing counterfactuals based on multiple distractor images. As a result, our method demonstrates superior overall performance.

    
  

\begin{figure*}[htbp]
  \centering
  \begin{subfigure}[b]{\textwidth}
    \centering
    \includegraphics[width=0.24\textwidth]{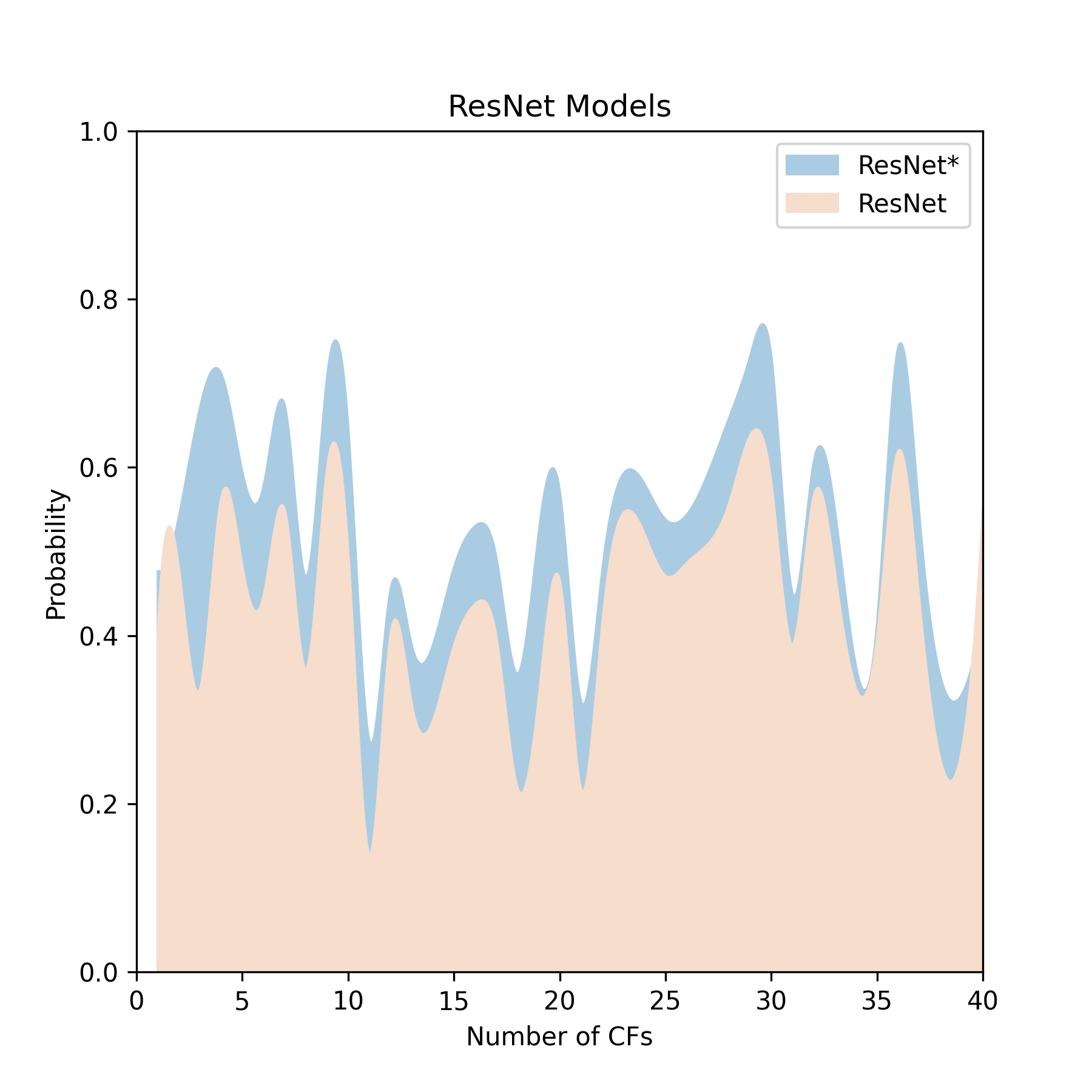}
    \includegraphics[width=0.24\textwidth]{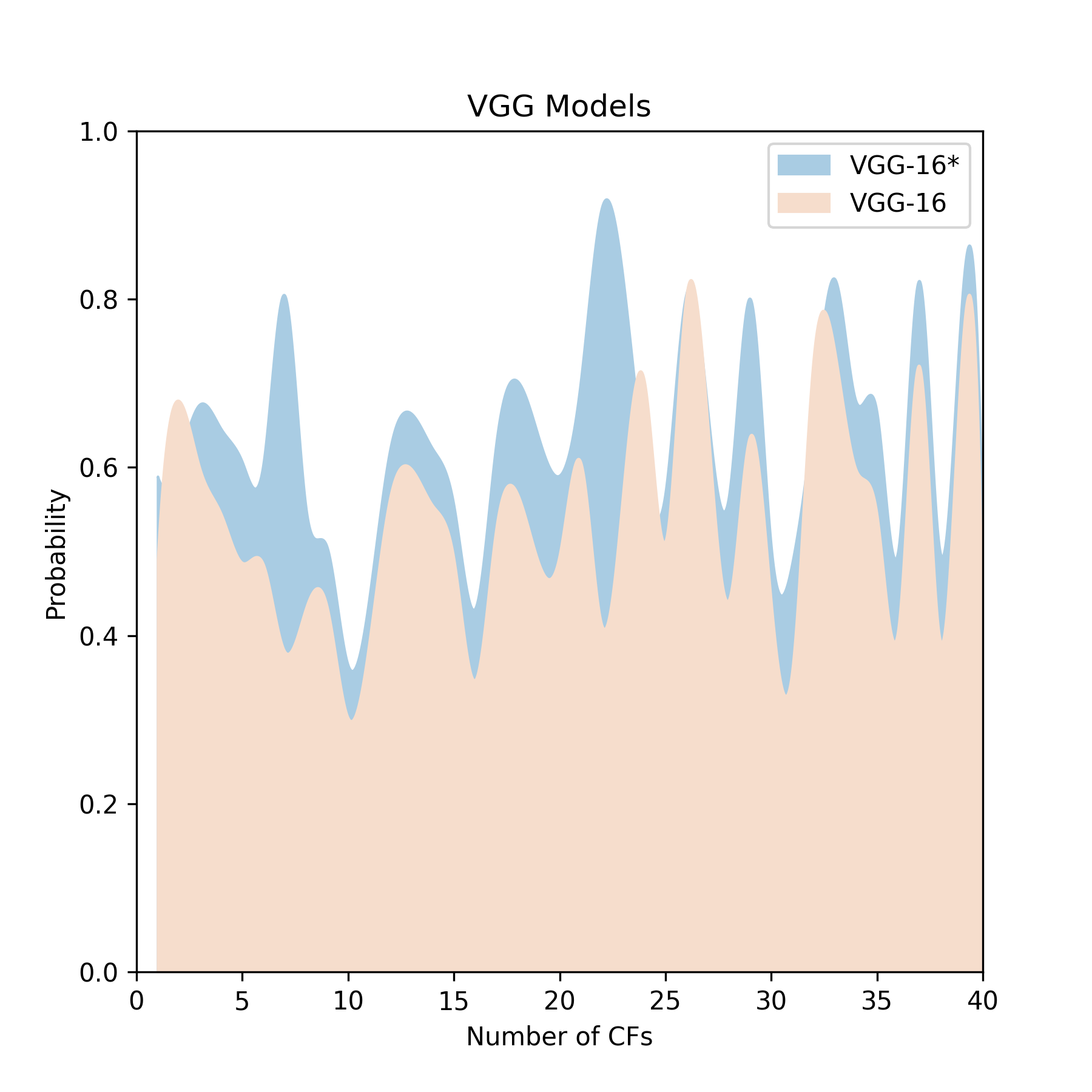}
    \includegraphics[width=0.24\textwidth]{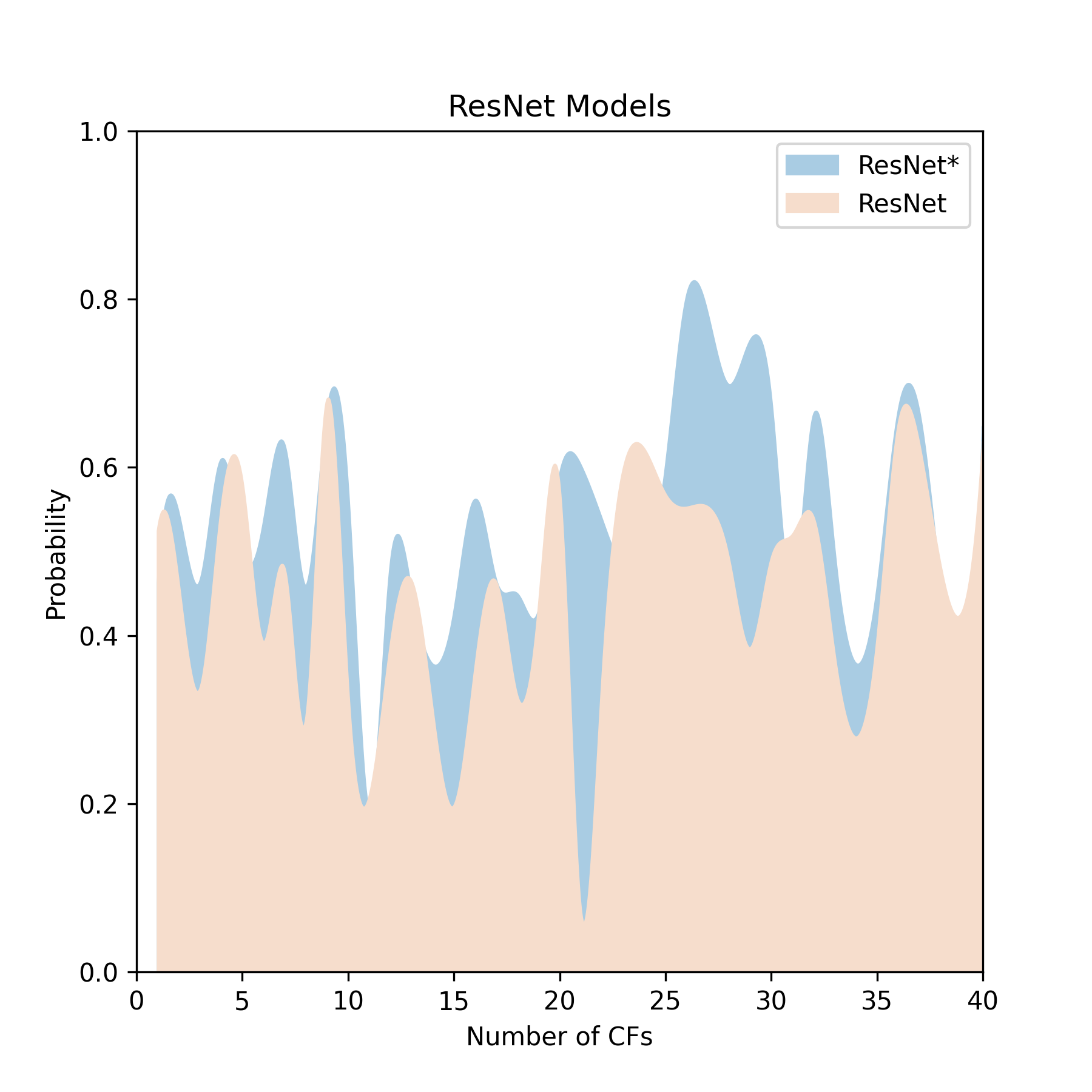}
    \includegraphics[width=0.24\textwidth]{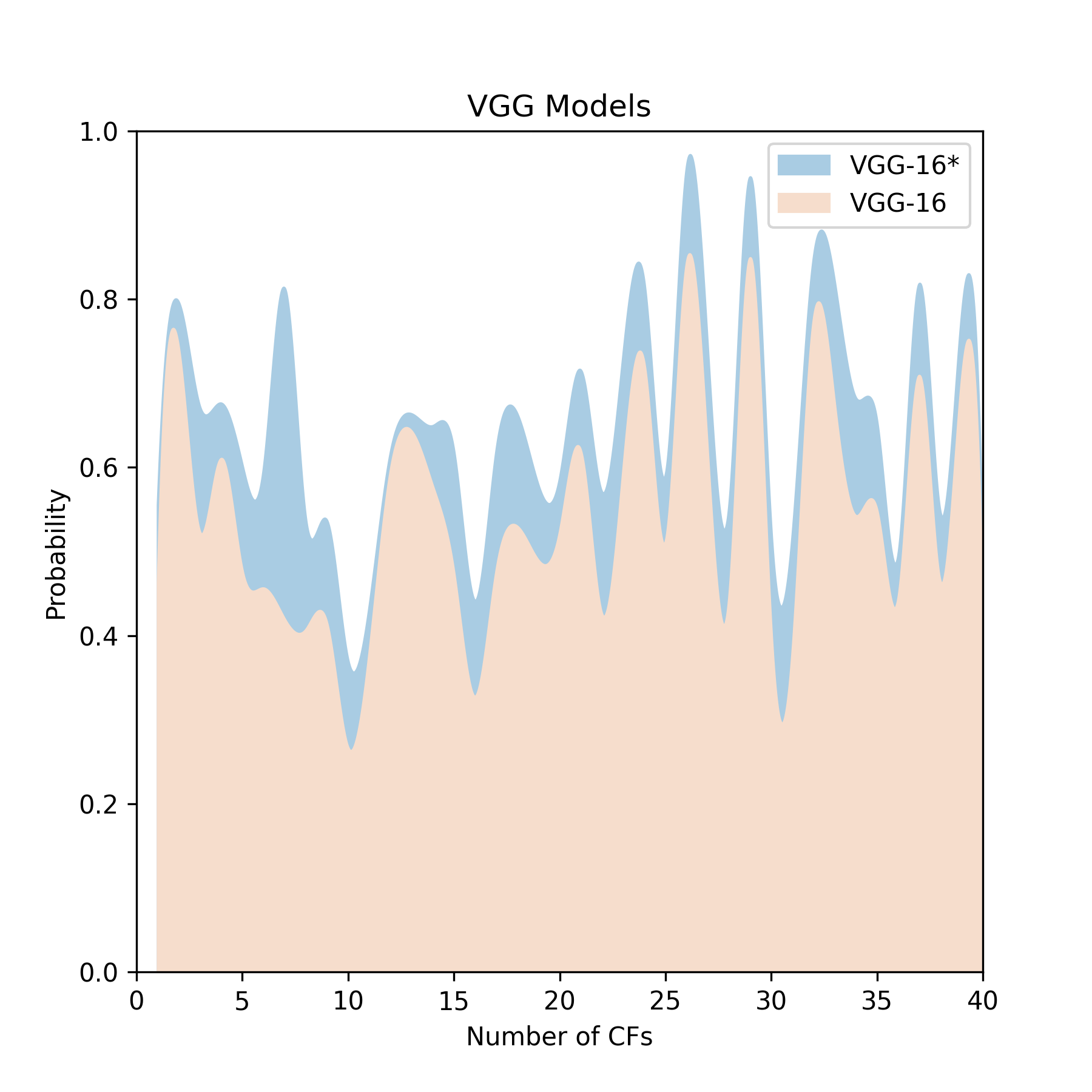}
    \caption{Assessment of class probabilities for each generated counterfactual}
    \label{fig:subfig1}
  \end{subfigure}
  \vspace{1em}
  \begin{subfigure}[b]{\textwidth} 
    \centering
    \includegraphics[width=0.24\textwidth]{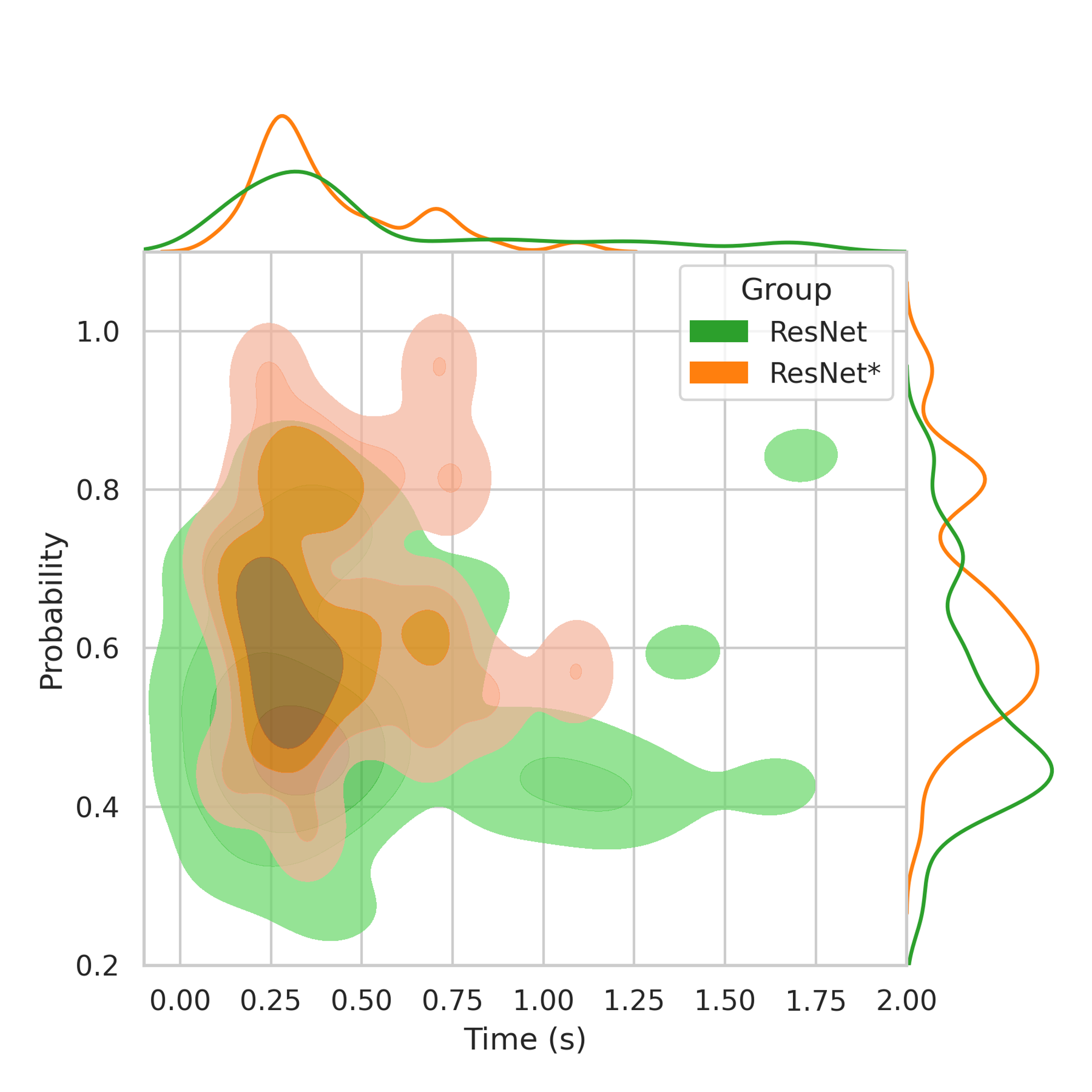}
    \includegraphics[width=0.24\textwidth]{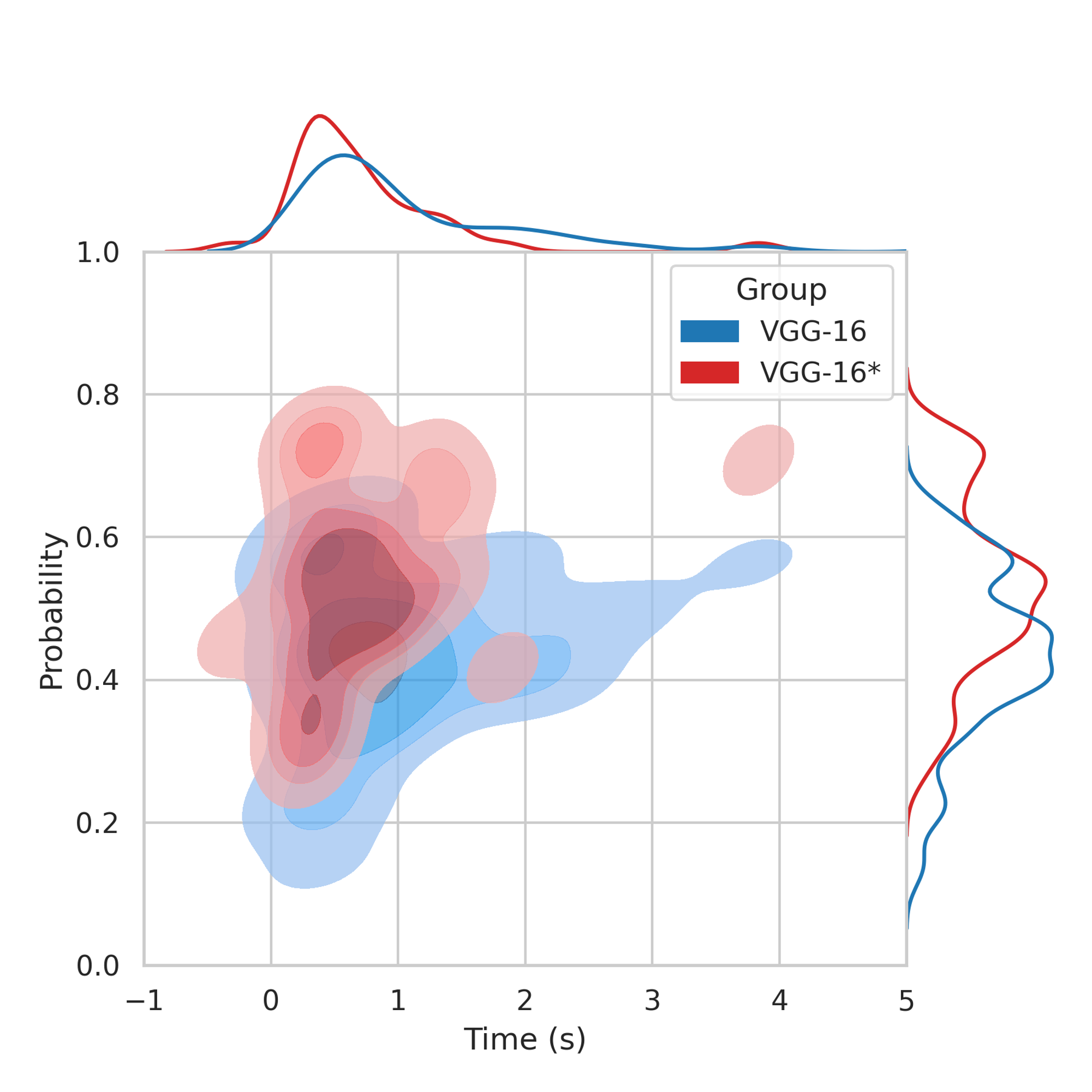}
    \includegraphics[width=0.24\textwidth]{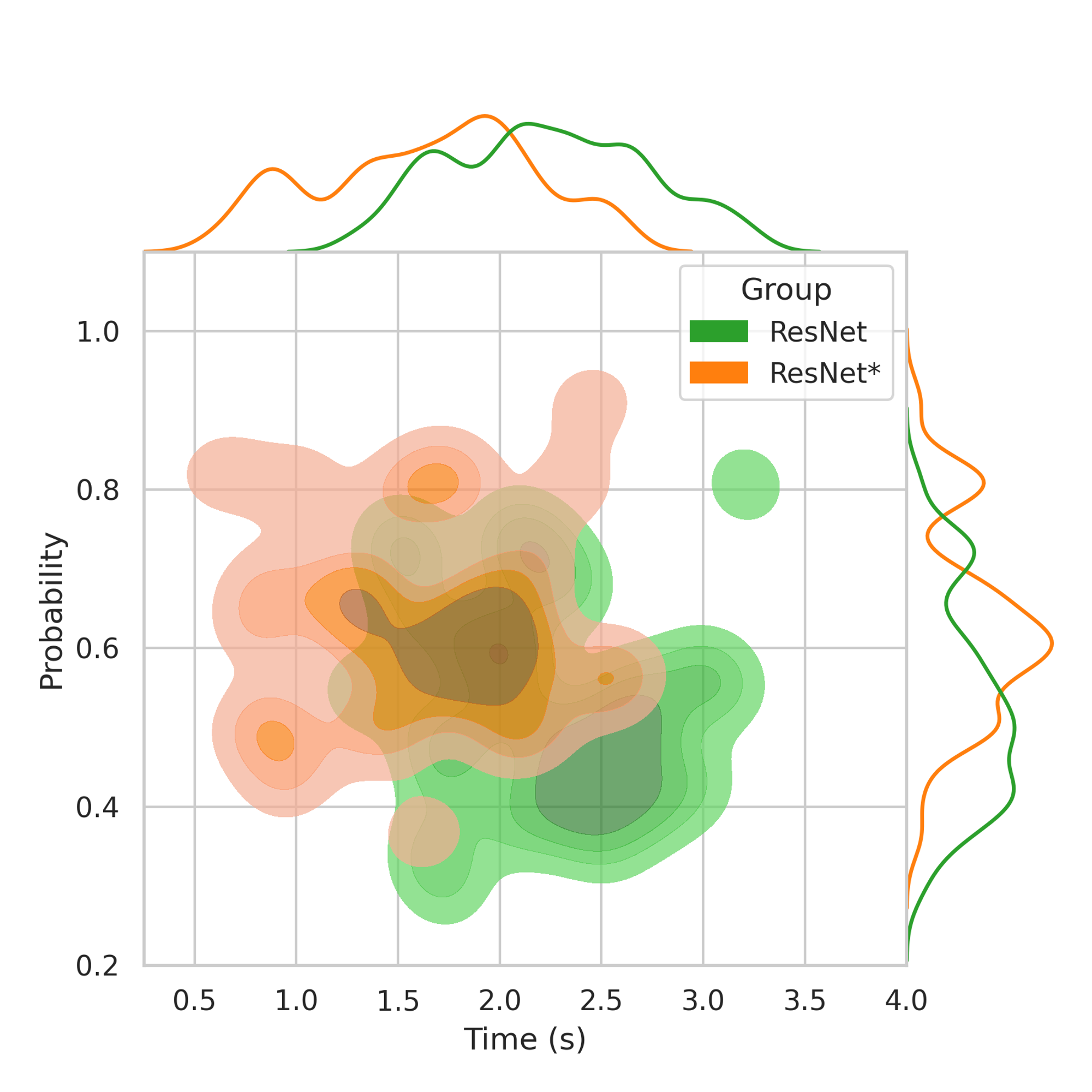}
    \includegraphics[width=0.24\textwidth]{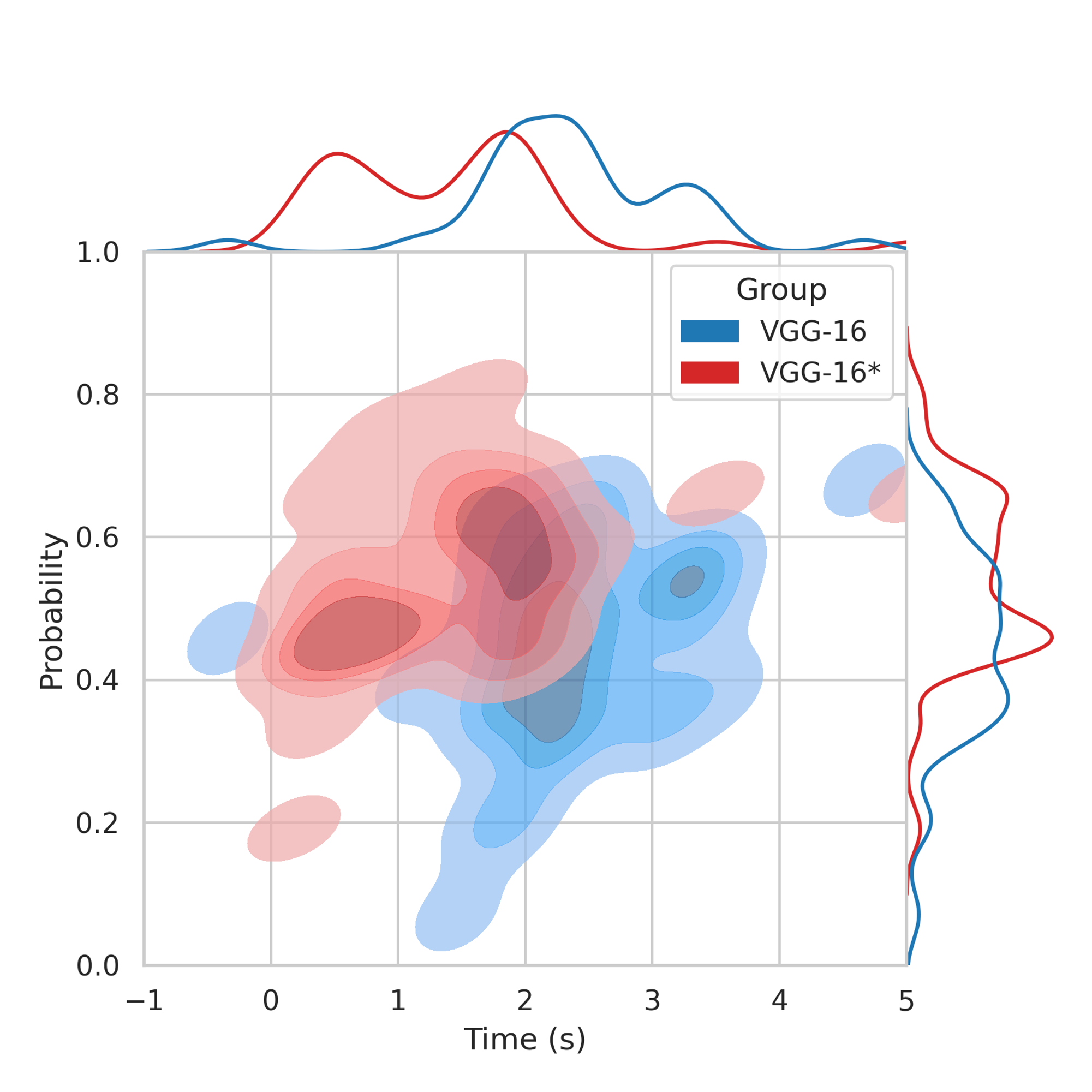}
    \caption{Efficiency-quality distribution for counterfactual computing}
    \label{fig:subfig2}
  \end{subfigure}
  \caption{Validation of efficiency for counterfactual generation. (a) Class probability comparison of randomly selected (40) counterfactuals with the baseline method during the generation process on two datasets with ResNet-50 and VGG-16. (b) Comparative analysis of the time distribution for generation and the distribution of class probabilities among randomly selected (40) counterfactuals relative to the baseline method on two datasets with ResNet-50 and VGG-16. $(*)$ is our method, result of CUB-200-2011 dataset on left two columns, right two columns for Stanford Dogs dataset.}
  \label{fig:probability difference}

\end{figure*}

In visualization, we compare the semantic relevance of counterfactual regions edited from a single distractor image and multiple distractor images between our method and previous methods. As illustrated in Fig.~\ref{app1} (a), when utilizing a single distractor image for counterfactual computation, our method selects a region from that image for replacement in the query image's editing region. Thereby facilitating a predicted class change with maximized semantic relevance. In the results from the CUB-200-2011 and Stanford Dogs datasets, for the first and second groups of counterfactuals (columns 1 to 4), compared to the baseline method~\cite{13}, our approach primarily involves editing within feature regions that are semantically related to the target object, thereby generating more comprehensible counterfactuals. In Fig.~\ref{app1} (b), we demonstrate two sets of counterfactuals computed from multiple distractor images across two datasets. Specifically, the editing process based on multiple distractor images involves selecting all feature units from several distractor images for computation. In the results, we edited counterfactuals based on five distractor images, selecting only the feature region most semantic relevance to the query's replacement region from each distractor, in accordance with~\cite{14}. The regions edited up to the fourth modification prior to the counterfactual class transition and the final editing areas upon the transition to the transformed counterfactual class are highlighted with yellow frames. Both during the counterfactual editing process and in the feature areas edited for the final transition to counterfactual class, the previous method~\cite{14} lacks constraints related to the target object's semantics, selecting editing regions based solely on conditions of semantic similarity. This approach results in counterfactuals that are difficult to comprehend from a human perspective. Our method not only maintains the semantic consistency established by prior research but also ensures that the edited feature regions are semantically relevant to the target object.

\subsubsection{Analysis}

To further validate the efficiency of our method in generating counterfactuals, we compared the probabilities of the generated counterfactual classes and the time distribution required for each counterfactual generation with those of previous methods. Furthermore, we investigated the underlying factors affecting the efficiency of counterfactual generation, which previous work had not thoroughly examined. We proposed a validation method named Average Probability Difference for Class Transition (APD) metric to elucidate the reasons for the acceleration and deceleration of counterfactual generation.

First, we conducted a comparative evaluation on the CUB-200-2011 and Stanford Dogs datasets using ResNet-50 and VGG-16 to assess the class probabilities and generation times of forty randomly selected counterfactuals during the counterfactual generation process, relative to baseline methods.  Fig.~\ref{fig:probability difference} (a) illustrates that the final counterfactual class probabilities calculated using our method are generally higher than those of the baseline. In Fig.~\ref{fig:probability difference} (b) illustrates a distribution chart showcasing the time required for each counterfactual edit alongside the associated counterfactual class probabilities. In the comparative analysis using ResNet-50 on the CUB-200-2011 dataset, compared to the baseline method, the distribution graphs of the probabilities and generation times for the counterfactual classes produced by our method demonstrate higher probabilities and are more concentrated within a narrower time frame.
The data reveal a trend wherein our method achieves higher counterfactual class probabilities with the least time consumption, thereby validating the efficiency of counterfactual generation.

\begin{figure}[htb]
    \centering
    \includegraphics[width=0.6\textwidth]{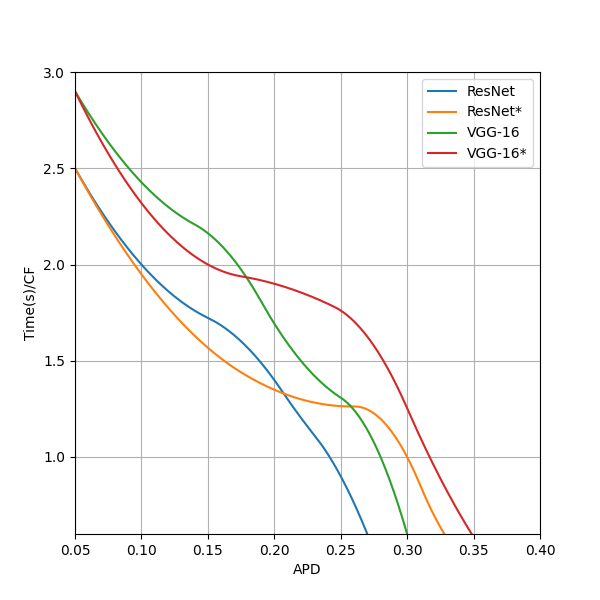}
    \caption{The curve of the APD metric versus the time required for each counterfactual generation, $*$ is our method.}
    \label{APD}
\end{figure}

Second, to further investigate the intrinsic reasons for the efficiency differences in counterfactual generation compared to previous methods, prior research~\cite{14} solely verified the differences in counterfactual generation efficiency by measuring the number of counterfactuals edited per second.  This approach clearly delineates the time differences in counterfactual computation between editing a single distractor image~\cite{13} and multiple distractor images~\cite{14}. However, it did not explore the underlying reasons for these time differences. To delve deeper into the causes of these discrepancies, we hypothesize that the efficiency of counterfactual editing depends on the magnitude of increase in the predicted probability of the counterfactual class with each edit. If the increment in the probability of the counterfactual class is substantial with each edit, it implies that fewer edits are needed to generate a counterfactual, thus reducing the time required for its generation.

Therefore, to elucidate the reasons underlying the differences in counterfactual efficiency, we propose a new metric : Average Probability Difference for Class Transition (APD). $\Delta P_k$ measures the magnitude of probability change of the $k^{th}$ edit, and the averaged $\Delta P$ is computed as follows:
    \begin{equation}
    \Delta P= \frac{1}{n} \sum_{k=1}^{n} \Delta P_k,
    \end{equation}
    where $n$ is the number of total edits for each counterfactual generation.

As illustrated in Fig.~\ref{APD}, to more clearly display the visualization results, we primarily extracted counterfactual samples within the most concentrated range of APD values (0.1 to 0.4) along with the corresponding generation times.  It is evident that our method generates more counterfactuals with higher APD values compared to the baseline, and the higher the APD, the less time is required to generate a counterfactual.

\begin{table*}[h]
\centering
\caption{Ablation study on weighted semantic map (WS), segmentation matrix (S), and auto-adaptive candidate editing sequence (AES) for prior works in ResNet-50 and VGG-16 model on CUB-200-2011 and Stanford Dogs datasets.}
\resizebox{\textwidth}{!}{ 
\begin{tabular}{l ccc|ccc|ccc|ccc}
\toprule
& \multicolumn{6}{c}{CUB-200-2011} & \multicolumn{6}{c}{Stanford Dogs Extra} \\
\cmidrule(lr){2-7} \cmidrule(lr){8-13}
& \multicolumn{3}{c}{ResNet-50} & \multicolumn{3}{c}{VGG-16} & \multicolumn{3}{c}{ResNet-50} & \multicolumn{3}{c}{VGG-16} \\
\cmidrule(lr){2-4} \cmidrule(lr){5-7} \cmidrule(lr){8-10} \cmidrule(lr){11-13}
Method & Near-KP & Same-KP & \# Edits & Near-KP & Same-KP & \# Edits & Near-KP & Same-KP & \# Edits & Near-KP & Same-KP & \# Edits \\ 
\midrule
Goyal~\cite{13} & 50.9 & 6.8 & 3.5 & 54.6 & 8.3 & 5.5 & 34.9 & 3.6 & 4.3 & 35.7 & 16.4 & 6.3  \\
Goyal~\cite{13}+\textbf{WS~(w/o S)}+\textbf{AES} & 53.9 & 7.0 & 3.5  & 57.5 & 8.8 & 5.2  & 36.2 & 4.0 & 4.3  & 37.1 & 17.2 & 6.2  \\
Goyal~\cite{13}+\textbf{WS} & 55.4 & 7.0 & 3.2  & 57.5 & 8.8 & 5.4 & 37.0 & 5.2 & 4.3 & 37.6 & 17.9 & \underline{6.0}  \\
Goyal~\cite{13}+\textbf{WS}+\textbf{AES} & 57.6 & 6.9 & 3.0  & 59.4 & 9.1 & 5.1 & 38.2 & 5.6 & \textbf{4.1} & 38.4 & \underline{18.6} & 5.8 \\
Vandenhende~\cite{14} & 60.3 & 30.2 & 3.2 & 68.5 & 35.3 & \underline{3.9} &37.2 & 16.7 & 4.8 &37.5 & 16.4 & 6.6 \\
Vandenhende~\cite{14}+\textbf{WS~(w/o S)}+\textbf{AES} & \underline{64.5} & \underline{31.3} & \underline{2.9} & \underline{72.0} & \underline{36.5} & 3.8  & 40.1 &18.4  & 4.6 & 39.1 & 17.2 & 6.4  \\
Vandenhende~\cite{14}+\textbf{WS} & 61.9 & 24.2 & 2.8  & 69.7 & 35.8 & 3.8  & \underline{41.3} & \underline{18.8} & 4.2  & \underline{39.4} & 17.9 & 6.3 \\
Vandenhende~\cite{14}+\textbf{WS}+\textbf{AES} & \textbf{64.9} & \textbf{31.4} & \textbf{2.8} & \textbf{73.1} & \textbf{36.7} & \textbf{3.8} & \textbf{43.5} & \textbf{19.3} & \underline{4.2} &\textbf{ 40.3} & \textbf{19.6} & \textbf{5.8} \\
\bottomrule
\end{tabular}
}
\label{ag}
\end{table*}

\subsubsection{Ablation Study} In this study, we conducted ablation studies in four primary experiments. The first experiment aimed to assess the impact of integrating our method with existing approaches. The second experiment involved evaluating the influence of including or excluding our proposed optimization selection function on the performance of the counterfactual generation process. The third experiment investigated whether the feature units of distractor images require ranking operations during the computation of counterfactuals and explored their impact on the generation of counterfactuals. The fourth experiment investigates the impact of the external model's performance on our method.

For assessing the impact of integrating our novel weighted semantic map and auto-adaptive candidate editing sequence into these existing frameworks, detailed in Table.~\ref{ag}. Additionally, we investigated how the presence or absence of a segmentation matrix used for semantic supervision within our proposed modules affects performance.  Results on the CUB-200-2011 dataset using ResNet-50 indicate that, after integrating our WS and AES modules with the baseline method~\cite{13}, the Near-KP metric improved by 6.7\% and the number of edits was reduced by 0.5.  In contrast, the latest studies~\cite{14} show that when our modules are combined, there's an uplift of 4.6\% and 1.2\% in the Near-KP and Same-KP metrics, respectively, alongside a decrement of 0.4 in the number of edits.

\begin{table}[ht]
\centering
\caption{Evaluating the impact of optimization selection function and loss function for performance of counterfactual generation.}
\resizebox{0.7\linewidth}{!}{
\begin{tabular}{@{}cccccccc@{}}
\toprule
 $L_\text{sim}$ & $L_\text{o}$ & Time (s) & Near-KP & Same-KP & \# Edits & APD & AT (m)\\ 
\midrule
 $\times$ & $\times$ & 2.2 &  55.2 & 22.1 & 3.4  & 0.08 & 63 \\
 \checkmark & $\times$ & 2.2&  60.3 & \underline{30.2} & 3.2  & 0.09 & 65 \\
 $\times$ & \checkmark & \textbf{1.7} &  \underline{62.7} & 22.5 & \underline{2.9}  & \underline{0.12} & \textbf{26} \\
 \checkmark & \checkmark & \underline{2.0} &  \textbf{64.9} & \textbf{31.1} & \textbf{2.8}  & \textbf{0.14}& \underline{32} \\
\bottomrule
\end{tabular}
}
\label{loss}
\end{table}

To validate the impact of including or excluding our proposed optimization selection function as well as loss functions introduced in prior work~\cite{14} on the performance of counterfactual generation, as illustrated in Table.~\ref{loss}, the result indicate that incorporating the similarity loss function $L_\text{sim}$ has enhanced semantic metrics, with Near-KP and Same-KP improving by 5.1\% and 10.1\%, respectively. Meanwhile, the implementation of the optimization selection function $L_\text{o}$ markedly improves the efficiency of generating each counterfactual, as evidenced by reductions in the time required and the number of edits along with improvements in the APD metric. Due to the efficiency of the $L_\text{o}$ in accelerating class transitions during counterfactual editing, it also reduces the all time (AT) consumed by the entire process. In general, the computational workflow of the method sequentially edits each query's counterfactual image, and once a class transition occurs, it immediately proceeds to edit the next counterfactual.

\begin{table}[ht]
\centering
\caption{The ablation study results on whether to perform the weighted semantic map (WS) and auto-adaptive candidate editing sequence (AES) process on distractor images for counterfactual editing with ResNet-50.}
\resizebox{0.7\linewidth}{!}{
\begin{tabular}{@{}cccccccc@{}}
\toprule
 WS & AES & \# CFs & Near-KP & Same-KP & Avg. prob. & \# Edits &AT (m) \\ 
\midrule
 $\times$ & $\times$ & 4,407 & 60.1 & \textbf{27.5} &67.3  & 3.2 & 63\\
 \checkmark & $\times$ & 4,407 & \textbf{60.5}~(+0.4) & 27.5&  \underline{68.4}~(+1.1) & \underline{3.1} & \underline{36} \\
 \checkmark & \checkmark & 3,577 &59.2~(-0.9) & \underline{25.4}~(-2.1) & \textbf{75.4}~(+8.1) & \textbf{2.8} & \textbf{23}\\

\bottomrule
\end{tabular}
}
\label{wsaes}
\end{table}

\begin{table*}[h]
\centering
\caption{ We randomly selected thirty query images with complex scenes and evaluated the impact of SAM-generated (SAM) low-quality masks and manually created ground truth (GT) masks on the performance of counterfactual generation.}
\resizebox{\textwidth}{!}{
\begin{tabular}{l ccc ccc ccc ccc}
\toprule
& \multicolumn{6}{c}{CUB-200-2011} & \multicolumn{6}{c}{Stanford Dogs Extra} \\
\cmidrule(lr){2-7} \cmidrule(lr){8-13}
& \multicolumn{3}{c}{ResNet-50} & \multicolumn{3}{c}{VGG-16} & \multicolumn{3}{c}{ResNet-50} & \multicolumn{3}{c}{VGG-16} \\
\cmidrule(lr){2-4} \cmidrule(lr){5-7} \cmidrule(lr){8-10} \cmidrule(lr){11-13}

\multicolumn{13}{c}{\textbf{~~~~~~~~~~~~~~~~~~~Single edit}} \\
\midrule
Method & Near-KP & Same-KP & \# Edits & Near-KP & Same-KP & \# Edits  & Near-KP & Same-KP & \# Edits  & Near-KP & Same-KP & \# Edits \\
\midrule
SAM~\cite{16} & 71.2 & 29.1 & -   & 73.6 & 37.4 & -  &\textbf{ 53.8} & 22.1 & -    &48.6 & 21.2 & - \\
GT & \textbf{75.1} & \textbf{29.3} & -  &\textbf{74.3} & \textbf{37.8} & - & 53.6 & \textbf{22.6} & -  & \textbf{49.8} & \textbf{21.8} & - \\

\cmidrule{1-13}
\multicolumn{13}{c}{\textbf{~~~~~~~~~~~~~~~~~~All edits}} \\
\midrule
SAM~\cite{16}  & 60.3 & 28.8 & 3.1 & 69.1 & 36.4 & 3.8  & 37.2 & 17.8 & 4.2  & 39.0 & 17.1 & 6.2 \\
GT  & \textbf{61.1} & \textbf{31.2} & \textbf{2.8 } & \textbf{70.4} & \textbf{36.7} & \textbf{3.6}  & \textbf{38.4} & \textbf{19.1} & \textbf{4.0}  & \textbf{39.6} & \textbf{17.8} & \textbf{5.6}  \\
\bottomrule
\end{tabular}
}
\label{SAMGT}
\end{table*}

In addition, our method focuses solely on generating a weighted semantic map for the query image and subsequently ranking it, without performing similar operations on the distractor images.  Therefore, we assessed the necessity of generating and ranking the weighted semantic map for the distractor images. As depicted in Table.~\ref{wsaes}, the WS module brings a modest 4\% boost in the Near-KP metric, attributed to the WS module's editing focus being confined to the semantic region identified within the segmentation map. This module avoids computing counterfactuals in non-semantic regions, thereby reducing the number of editing combinations and influencing the required number of edits (\# Edits) for generating counterfactuals.

On the other hand, utilizing the WS and AES modules together on distractor images favors replacing feature units crucial to the distractor class into the query image, while these modules accelerate the class transition of the query, boosting the average probability of the counterfactual class by 8.1\%. However, there was a performance decline across metrics related to semantics, and the total number of generated counterfactuals also decreased. Correspondingly, the AT consumed by the overall process is also significantly reduced. This is because it is particularly challenging to identify editing combinations that are semantically similar and crucial for their respective classes within the semantic regions of both the query and distractor images. This process filters out a large number of potential editing combinations and results in query images that cannot be edited into counterfactuals. Therefore, we only consider computations within the semantic region of the distractor images, rather than proceeding with further sorting operations.

Finally, to explore the impact of the external model on our method, we selected 30 low-quality query masks generated by SAM in complex scenarios(filtered based on the Intersection Over Union metric) and compared the performance of our method using these masks to that using manually created ground truth masks. As shown in Table.~\ref{SAMGT},while manually created high-quality masks further improve overall performance, even in the extreme case of using low-quality masks, the performance does not degrade significantly compared to the baseline. Although the mask quality determines whether irrelevant feature units are involved in the counterfactual process, potentially filtering out some relevant units, the optimal editing sequence in our method ensures that editing always begins with the most important feature units for class decision-making. Consequently, the number of editing steps is not significantly affected. Additionally, since the most similar unit combinations are consistently used for calculation throughout this process, the Same-KP metric remains largely unaffected.


\begin{figure}[htbp]
  \centering
  \begin{subfigure}[b]{0.48\textwidth}
    \centering
    \includegraphics[width=\textwidth]{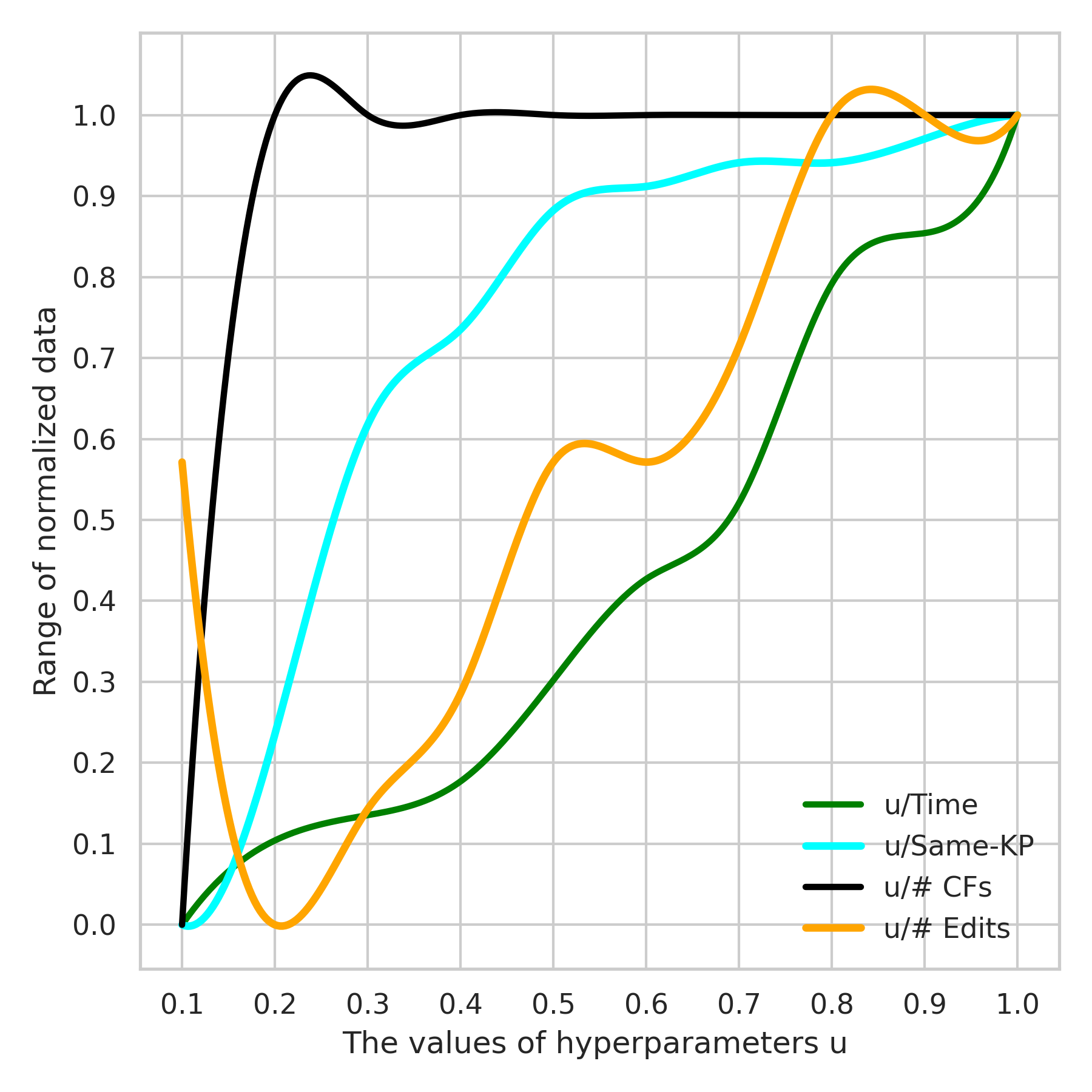}
    \caption{The impact of varying values of the threshold $u$}
    \label{fig:subfig_ut1}
  \end{subfigure}
  \hfill 
  \begin{subfigure}[b]{0.48\textwidth}
    \centering
    \includegraphics[width=\textwidth]{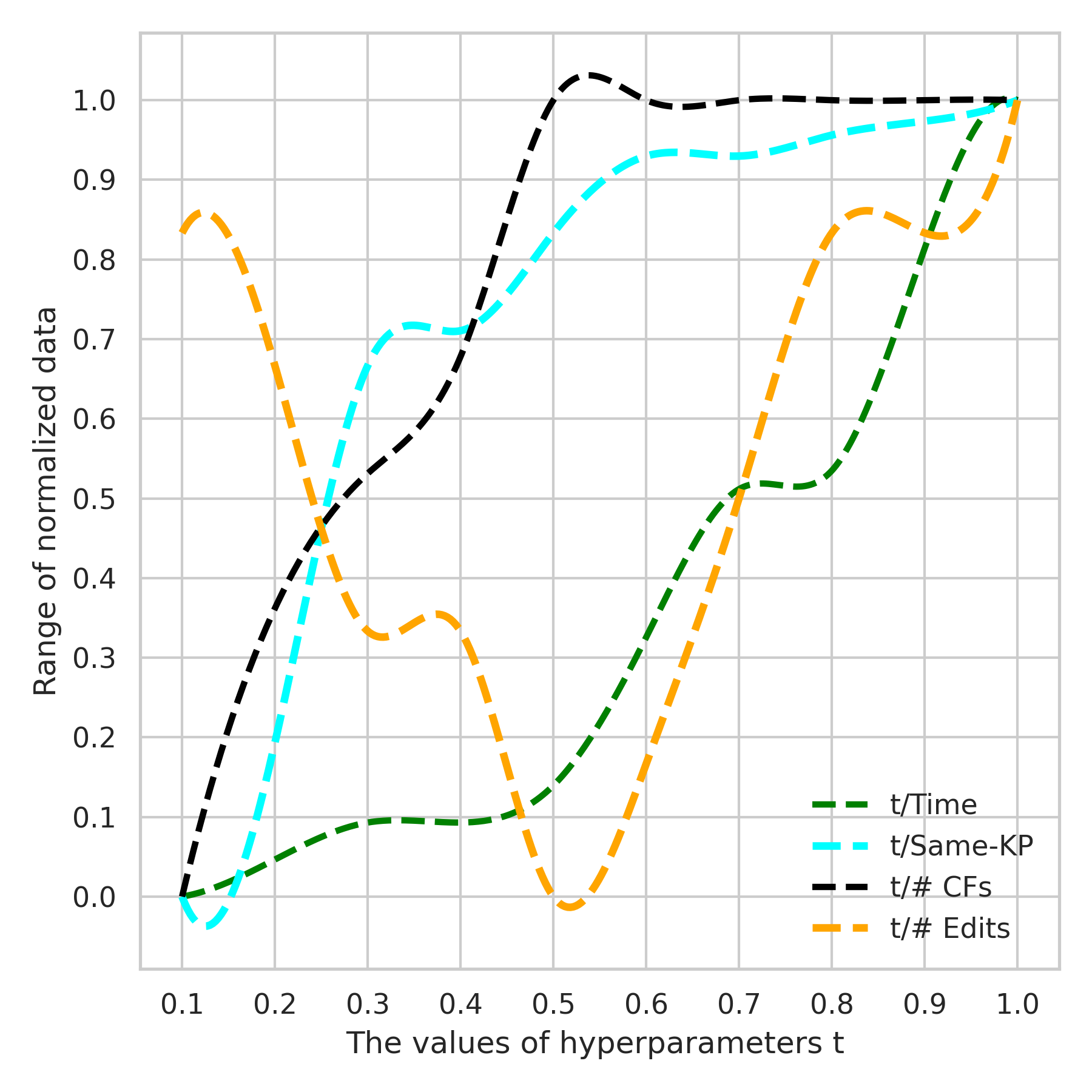} 
    \caption{The impact of varying values of the threshold $t$}
    \label{fig:subfig_ut2}
  \end{subfigure}
  
  \caption{The impact of different values for the similarity threshold $u$ and weight score threshold $t$ on counterfactual generation.}
  \label{figut}
\end{figure}

\begin{table}[ht]
\centering
\caption{Performance metrics at various values of the similarity threshold $u$ and the weight score threshold $t$ during the counterfactual editing process.}
\caption*{\textbf{ For similarity threshold $u$}}
\resizebox{0.5\linewidth}{!}{
\begin{tabular}{@{}cccccc@{}}
\toprule
 $u$ & Time (s)  & Same-KP  & \# CFs & \# Edits \\ 
\midrule
 1.0  & 10.8 & 33.5 & 4,407 & 3.5\\
 0.9  & 9.4 & 33.4 & 4,407 & 3.5\\
 0.8  & 8.8 & 33.4 & 4,407 & 3.5\\
 0.7  & 6.2 & 33.4 & 4,407 & 3.3\\
 0.6  & 5.3 & 33.3 & 4,407 & 3.2\\
 0.5  & 4.1 & 33.1 & 4,407 & 3.2\\
 0.4  & 2.9 & 32.6 & 4,407 & 3.1\\
 0.3  & 2.5 & 32.2 & 4,407 & 3.0\\
 0.2  & 2.2 & 30.1 & 4,407 & 2.9\\
 0.1  & 1.2 & 30.1 & 4,146 & 3.2\\
\bottomrule
\end{tabular}
}
\bigskip 
\caption*{\textbf{ For weight score threshold $t$}}
\resizebox{0.5\linewidth}{!}{
\begin{tabular}{@{}cccccc@{}}
\toprule
 $t$ & Time (s)  & Same-KP  & \# CFs & \# Edits \\ 
\midrule
 1.0  & 6.1 & 33.5 & 4,407 & 3.5\\
 0.9  & 5.3 & 33.2 & 4,407 & 3.4\\
 0.8  & 4.1 & 33.0 & 4,407 & 3.4\\
 0.7  & 4.0 & 32.7 & 4,407 & 3.2\\
 0.6  & 3.2 & 32.7 & 4,407 & 3.5\\
 0.5  & 2.4 & 31.6 & 4,407 & 2.9\\
 0.4  & 2.2 & 30.2 & 4,026 & 3.1\\
 0.3  & 2.2 & 29.7 & 3,677 & 3.1\\
 0.2  & 2.0 & 24.3 & 3,274 & 3.3\\
 0.1  & 1.8 & 22.1 & 2,413 & 3.4\\
\bottomrule
\end{tabular}
}
\label{u&t}
\end{table}

\subsubsection{Sensitivity}

To validate the impact of varying values of the similarity threshold $u$ used after computing  semantic consistency loss and the weight score threshold $t$ employed during the weighted score ranking process on the performance within the counterfactual computation process,  
Table.~\ref{u&t} illustrates the impact of different thresholds $u$ and $t$ on performance results. We have set the increment for these two threshold variations at 0.1. When either threshold $u$ or $t$ undergoes a change, the other threshold is fixed at 1 and remains unchanged.
The findings reveal that setting $u$ at 0.2 maximizes the time to generate each counterfactual and optimizes metrics such as Same-KP and the average number of edits. Focusing solely on units with the highest similarity to select smaller values can lead to the elimination of certain editing combinations that might be beneficial for class transitions, consequently resulting in a reduction of 261 in the total number of generated counterfactuals. Conversely, a $t$ threshold of 0.5 yields improved overall performance metrics. Setting the threshold to 0.1 during the ranking process results in prioritizing feature units with the lowest weight scores. This adjustment increases the average number of edits necessary for generating counterfactuals and substantially filters out many feasible editing combinations that could facilitate class transitions, thereby halving the total number of generated counterfactuals. To visually present the outcomes detailed in the aforementioned table, Fig.~\ref{figut} depicts the variation curves of different metrics as the values of $u$ and $t$ rise. Given the substantial variance in the data represented by each metric, we preemptively normalized the data for each metric to ensure comparability. In summary, after evaluating the performance impact, we select the optimal threshold values of $u$ at 0.2 and $t$ at 0.5.

\section{Conclusion}
In this study, we propose that the weighted semantic map phase ensures the semantic relevance of the edited feature regions to the target object during the counterfactual editing process.  The auto-adaptive candidate editing phase significantly enhances the efficiency of counterfactual class transitions, which is particularly crucial for non-generative visual counterfactual explanation methods.

Nonetheless, we further minimize the required editing region by utilizing masks obtained from segmentation models. While this approach provides valuable semantic context, it is computationally intensive, especially when applied to large datasets. Moreover, even the limited use of external segmentation models during the preprocessing phase may reduce the interpretability of our counterfactual editing method, as it adds to the overall complexity of the approach.

In future work, our goal is to derive weighted semantic maps directly from images through learning, utilizing internal semantics rather than relying on external segmentation models for guidance. This approach aims to simplify the counterfactual editing process and enhance the model’s interpretability. We anticipate developing a more intuitive and effective method to generate meaningful fine-grained counterfactual visual explanations.

\section*{Acknowledgement}
We would like to express our sincere gratitude to Hayang Jo for his valuable contributions to this work, particularly in refining the formulas and ensuring adherence to writing standards. This work was supported by Institute of Information \& communications Technology Planning \& Evaluation (IITP) grant funded by the Korea government (MSIT) (No. 2022-0-00984, Development of Artificial Intelligence Technology for Personalized Plug-and-Play Explanation and Verification of Explanation \& No. RS-2019-II190079, Artifcial Intelligence Graduate School Program, Korea University)

\end{document}